\documentclass[11pt,oneside,letterpaper]{article}
              
\newif\ifaistats
\aistatsfalse

\usepackage{Macros} 

\begin{document}

\title{High-Probability Bounds For Heterogeneous Local Differential Privacy}
\author{Maryam Aliakbarpour\footnote{Department of Computer Science and Ken Kennedy Institute} \\
  Rice University \\
  \texttt{maryama@rice.edu} \\
  \and
  Alireza Fallah\footnotemark[1] \\
  Rice University \\
  \texttt{afallah@rice.edu} \\
  \and
  Swaha Roy \\
  Rice University \\
  \texttt{sr112@rice.edu} \\
  \and
  Ria Stevens \\
  Rice University \\
  \texttt{ria.stevens@rice.edu}
}

\date{October 14, 2025}
\maketitle

\sloppy

\begin{abstract}
    We study statistical estimation under local differential privacy (LDP) when users may hold heterogeneous privacy levels and accuracy must be guaranteed with high probability. Departing from the common in-expectation analyses, and for one-dimensional and multi-dimensional mean estimation problems, we develop finite sample upper bounds in $\ell_2$-norm that hold with probability at least $1-\beta$. We complement these results with matching minimax lower bounds, establishing the optimality (up to constants) of our guarantees in the heterogeneous LDP regime. We further study distribution learning in $\ell_\infty$-distance, designing an algorithm with high-probability guarantees under heterogeneous privacy demands. Our techniques offer principled guidance for designing mechanisms in settings with user-specific privacy levels.
\end{abstract}



\clearpage




\section{Introduction}
The unprecedented growth of data collection has made protecting user privacy a central challenge.
Local Differential Privacy (LDP) offers a compelling solution, enabling the analysis of population-level statistics without exposing any individual’s raw data—even to the aggregator~\citep{Dwork2006, Kasiviswanathan2011}. In this model, each user perturbs their own data before transmission, ensuring that only randomized reports reach the curator. This paradigm has moved well beyond theory, with large-scale deployments at Google~\citep{erlingsson2014rappor}, Microsoft~\citep{ding2017collecting}, and Apple~\citep{apple2017learning, thakurta2017learning}.

Most research on LDP, however, rests on two simplifying assumptions: that all users share a uniform privacy guarantee ($\varepsilon$) and that error bounds only need to hold in expectation or just with a constant probability. In this work, we move beyond this idealized model to study two crucial and more realistic variants of LDP.

The first variant is the \textit{heterogeneous privacy setting}, where each of the $n$ users may have their own distinct privacy parameter $\varepsilon_i$. This framework acknowledges that in practice, populations are diverse: some users may demand stronger protections, while others are willing to trade privacy for utility. While tuning local perturbation protocols to different $\varepsilon_i$ is straightforward, accurately aggregating these heterogeneous reports into global estimates poses significant algorithmic challenges.

The second, and complementary, variant concerns the need for \textit{high-probability results}—that is, error guarantees that hold with probability at least $1-\beta$. The conventional approach to achieving such results is to apply a standard amplification technique, like the median-of-means method, to an algorithm with in-expectation guarantees (e.g., Proposition 9 in~\citep{HsuS16}). However, this off-the-shelf approach suffers from two critical drawbacks:

\begin{enumerate}[leftmargin=*]
\item  \textit{Suboptimal Dependence on Confidence:} These generic methods yield a poor dependence on the confidence parameter $\beta$. They typically multiply the overall error by a factor polynomial in $\log(1/\beta)$, whereas an ideal method's error would increase only by an additive term involving $\log(1/\beta)$.

\item \textit{Breakdown of Repetition in Heterogeneous Settings:} These amplification techniques rely on a divide-and-conquer strategy, partitioning the population into subgroups to generate independent estimates. This strategy fails in the heterogeneous setting because it is non-trivial to ensure each subgroup has a statistically similar composition of privacy budgets. For example, if a few users have very large $\varepsilon_i$ values, they cannot be distributed evenly. The estimates from the subgroups containing them will be fundamentally different from the others, thereby breaking the statistical symmetry required for the median-of-means approach to work.
\end{enumerate}

While some prior work has studied heterogeneous LDP~\citep{fallah2024optimal, chaudhuri2023free, chaudhuri2023demands}, its error bounds only hold in expectation. For the very reasons just described—particularly the breakdown of repetition-based approaches—these guarantees cannot be readily converted into high-probability results, leaving the challenge of achieving robust, high-probability guarantees in this setting unaddressed.

In this work, we initiate the first systematic study that tackles these challenges simultaneously. We develop methods for heterogeneous LDP that provide tight high-probability error bounds without resorting to computationally expensive or statistically loose amplification techniques. Specifically, for both \textit{single- and multi-dimensional mean estimation}, we establish sharp characterizations of the trade-off between heterogeneous privacy budgets and estimation error. We design computationally efficient algorithms and prove matching lower bounds, demonstrating that our results are optimal up to constant factors. Moreover, our bounds achieve the desired additive logarithmic dependence on the confidence parameter $\failureProb$. Notably, for the high-dimensional LDP mean estimation problem, to our knowledge, no prior analysis has established optimal dependence on all parameters---even in the standard uniform- privacy setting.

Furthermore, as a key application of our results, we use our single-dimension mean estimator to address the problem of {\em distribution learning} in $\ell_\infty$-distance. We apply our techniques to the framework of~\cite{Bassily2015}, which relies on multiple mean estimation subroutines that must all be correct with high probability. Our high-probability estimation techniques naturally extend to this setting, yielding a new upper bound for distribution learning under heterogeneous local privacy.





\subsection{Problem Formulation}\label{sec:problem_formulation}
We begin by defining local differential privacy: 
\begin{definition}[Pure Local Differential Privacy (LDP) \citep{Kasiviswanathan2011}]
An algorithm $\mathcal{A} : \mathcal{X} \to \mathcal{Y}$ is said to be $\epsDP$-locally differentially private if for any measurable set $\mathcal{W} \subseteq \mathcal{Y}$ and every pair of points $X, X' \in \mathcal{X}$,
\begin{equation}
\PB{\mathcal{C}(X) \in \mathcal{W}} \leq e^{\epsDP} \; \PB{\mathcal{C}(X') \in \mathcal{W}}.
\end{equation}

\end{definition}

Consider a setting with $n$ users, each holding a data point $X_i$ drawn i.i.d.\ from an unknown distribution $P$ with parameter $\theta = \theta(P)$.  
Each user specifies a privacy parameter $\varepsilon_i$, indicating the level of protection they require. We denote the vector of these privacy parameters by $\eiVec \coloneqq \eis$.  
They transmit a perturbed report $Y_i$, which is an $\varepsilon_i$-locally differentially private version of $X_i$.  
The data curator receives $\{Y_i\}_{i=1}^n$ and computes an estimator $\hat{\theta}$ of $\theta$.  

Our goal is to design a protocol for estimating $\theta$ that respects each user’s privacy constraint.  
Formally, we construct $n$ privacy channels $\{\mathcal{C}_i\}_{i=1}^n$, where each  
$\mathcal{C}_i : \mathcal{X}_i \to \mathcal{Y}_i$ satisfies $\varepsilon_i$-LDP with respect to user $i$’s data.

We measure the quality of our estimator by finding an upper bound on the error of the estimation that holds with probability $1-\beta$. In particular, for any given $\beta \in (0,1)$, we are looking for the smallest $t\geq0$ for which: 
\begin{equation} \label{eqn:concentration_bound}
\PB{ \text{err}\left( \hat{\theta}(\bm{\mathcal{C}}(\bm{X})) - \theta \right) \leq t } \geq 1- \beta,
\end{equation}
where $\bm{\mathcal{C}}(\bm{X}) := \left( \mathcal{C}_1(X_1), \ldots, \mathcal{C}_n(X_n) \right)$.
We examine three specific variations of this problem. In each problem, the heterogeneous set-up remains the same, but the nature of the underlying distribution or the parameter may be different.
\paragraph{Single-Dimensional Mean Estimation:}
Each user $i \in [n]$ holds a data point $X_i$ drawn from a distribution $P$ with mean $\estParam$ over a bounded single-dimensional domain (either $[-1, +1]$ or $\{-1, +1\}$). The server estimates the mean of $P$, by aggregating each user's privatized data and outputting $\eiVec$-locally differentially private estimator $\hat{\estParam}$.


\paragraph{Multi-Dimensional Mean Estimation:}
Each user $i \in [n]$ holds a data point in the $\ell_2$ (Euclidean) ball of radius $\radius$, $X_i \in \mathbb{B}^{\dimValues}\left(\radius \right)$, drawn from a distribution $P$ with mean $\estParam$. The server estimates the mean of $P$, by aggregating each user's data and outputting $\eiVec$-locally differentially private estimator $\hat{\estParam}$.

\paragraph{Distribution Learning:}
For a fixed discrete distribution $p$ over $[d]$, each user $i \in [n]$ holds an item $x_i \in [d]$. We aim to estimate $\hat{p}(v)$, the estimator of the true proportion $p(v)$ of users holding item $v$ in $p$.

\subsection{Our Contributions}
\paragraph{Single-Dimensional Mean Estimation.}
We study mean estimation for single-dimensional bounded random variables, $X_i \in [-1,+1]$. Our core contribution in this setting is an \textit{optimized weighted average} of $Y_i$'s, where weights are determined by each user's privacy parameter, $\epsilon_i$. This method improves accuracy by down-weighting noisier contributions from users with stronger privacy requirements (smaller $\epsilon_i$), since their privatized data points are more distorted. Our approach leads to the following high-probability error bound.

\begin{theorem}[Informal version of Theorem~\ref{thm:laplace_mech} and Theorem~\ref{thm:RR}]
Let $\eiVec = \eis$ with $\ei \leq 1$ for all $i$ and $\failureProb \in (0, 1)$. Let $\PP$ be the family of distributions $P$ such that for any $X \sim P$, $X$ is bounded almost surely. Then, for all  $P \in \PP$, there exists an $\eiVec$-locally differentially private estimator $\hat{\estParam}$ such that with probability at least $1 - \failureProb$,
      \begin{equation} 
           \abs{ \hat{\estParam}\left( X_{1:n} \right) - \estParam \left( P \right) }^2 \leq \OO \p{ \min \left( \frac{\log \left( 1 / \failureProb \right)}{\sum_{i=1}^n \ei^2} , 1 \right) }.
      \end{equation}
\end{theorem} 

We show this bound is tight in Theorem~\ref{thm:lowerBound} using a high-probability instance of Le Cam's Lemma \citep{Ma2024}. To establish the upper bound, we provide two distinct algorithms. For general bounded distributions, Theorem~\ref{thm:laplace_mech} provides a formal guarantee for an estimator (Algorithm~\ref{alg:lap}) based on the \textit{Laplace mechanism}. As this mechanism requires communicating continuous values, which can be inefficient, we also present a communication-optimal method for binary data. In this setting, Theorem~\ref{thm:RR} proves the same tight bound is achievable with an estimator (Algorithm~\ref{alg:RR}) based on \textit{Randomized Response}.

\paragraph{Multi-Dimensional Mean Estimation.}
Our second main contribution is to extend our analysis to the \textit{multi-dimensional} setting, where we provide the first tight, high-probability bounds for mean estimation over the Euclidean ball under heterogeneous Local Differential Privacy (LDP).

\begin{theorem}[Informal version of Theorem~\ref{thm:highDimUB}]
Let $\eiVec = \eis$ where $\ei \leq 1$ for all $i$ and $\failureProb \in (0, 1)$. For all $P \in \PP_{2, \radius}$ such that for any $X \sim P$, $X \in \mathbb{B}^{\dimValues}\left(\radius \right)$ almost surely, there exists an $\eiVec$-locally differentially private estimator $\hat{\estParam}$ such that with probability at least $1 - \failureProb$,
\begin{equation}
    \norm{ \hat{\estParam}\left( X_{1:n} \right) - \estParam \left( P \right) }_2^2 \leq \OO \p{ \frac{\radius^2 \left( \dimValues + \log \left( 1 / \failureProb \right)\right)}{\sum_{i=1}^n \ei^2} }.  
\end{equation}
\end{theorem}
We present an efficient algorithm (Algorithm~\ref{alg:highDim}) that, similar to our one-dimensional estimator, computes an \textit{optimized weighted average} of privatized versions user data. Each user privatizes their high-dimensional data point using a mechanism of~\cite{duchi2013local}, which projects data points onto one of two hemispheres based on a randomized response selection. 
Our algorithm weights each received signal based on its sender's privacy level, $\epsilon_i$, to construct the final estimate.

The key novelty of our work lies in the high-probability nature of our bound. While bounds on the \textit{expected} error were known, converting them to high-probability guarantees is not straightforward, with standard techniques yielding suboptimal error rates. Our main technical contribution is a refined analysis to overcome this challenge. 
We exploit
the concentration of measure phenomenon that arises due to the privacy mechanism taking the form of a mixture of uniform distributions over hemispheres.
We achieve a bound with an optimal \textit{additive} dependence on the dimension and the confidence parameter (i.e., $d + \log(1/\delta)$), a significant improvement over the suboptimal multiplicative dependence ($d \cdot \log(1/\delta)$) that arises from standard approaches.

Finally, in Theorem~\ref{thm:highDimLB}, we prove that our algorithm is optimal by establishing a \textit{matching lower bound} via combination of Assouad and the high probability version of Le Cam method presented in~\cite{Ma2024}. This result solidifies our upper bound and provides a complete characterization of the high-probability error for this fundamental high-dimensional estimation problem.

\paragraph{Distribution Learning.} 
Our final contribution addresses distribution learning under heterogeneous local privacy.  
We adapt the projection-based frequency estimation algorithm of \cite{Bassily2015}, originally designed for the homogeneous setting, to distribution learning in the heterogeneous case.  
At a high level, our algorithm applies a Johnson–Lindenstrauss (JL) transform to compress user data into a lower-dimensional representation, privatizes a randomly selected coordinate via randomized response, and aggregates the resulting reports with carefully chosen weights to account for heterogeneous privacy levels.  
This yields a communication- and computation-efficient oracle to estimate the probabilities of the domain elements that enables distribution learning with $\ell_\infty$ error guarantees. 
\begin{theorem}[Informal version of Theorem~\ref{thm:HistEstBassilyUB}]
    Let $\eiVec = \eis$ where $\ei \leq 1$ for all $i$. Let $\PP$ be the family of distributions $P$ over $[d]$. Then, for all $P \in \PP$, there exists an $\eiVec$-locally differentially private estimator $\hat{P}$ such that
    \begin{equation}
       \norm{\hat{P}\left( \obsValue_{1:n} \right) - P}_\infty \leq \OO \left(\sqrt{ \frac{\log \left(\dimValues \right)}{\sum_{i=1}^n \ei^2}} \right) 
    \end{equation}
\end{theorem}
This theorem shows that our method achieves optimal dependence on the mixed privacy budgets, matching the rates known for the homogeneous case.

\subsection{Minimax Quantiles}\label{sec:minimaxQuantiles}

Standard techniques for lower bounding the expected error of an algorithm fail to capture the behaviour of its tail and the dependence on $\failureProb$. 
Therefore, to establish our lower bounds, we must rely on minimax quantiles \citep{Ma2024}.

\begin{definition}[LDP-Minimax Quantiles] 
\label{def:minimax_quantile}
    Let $\eiVec = \eis$, with $\ei \leq 1$ for all $i$. Let $\failureProb \in (0, 1)$. Let $\PP$ be a family of distributions, where $P_\theta \in \PP$ is parametrized by $\theta \in \estParamClass$. 
    Let $\QQ_{\eiVec}$ be the set of all conditional distributions $Q: \XX^{\otimes n} \rightarrow \YY^{\otimes n}$ guaranteeing $\eiVec$-local differential privacy.
    Let $\hat{\estParamClass}$ be the set of all measurable functions $\hat{\estParam} : \YY^{\otimes n} \rightarrow \estParamClass$. 
    Then, we define the minimax quantile, $\minimaxQ{\failureProb, \PP, \eiVec}$ as:
    \ifaistats
    \begin{align}
        &\minimaxQ{\failureProb, \PP, \eiVec} \coloneqq \underset{Q \in \QQ_{\eiVec}}{\inf} \; \underset{\hat{\estParam} \in \hat{\estParamClass}}{\inf} \; \underset{P \in \PP}{\sup} \; \inf \left\{ \minimaxThres \in [0, \infty) \; : \; \right. \nonumber \\
        &\left. \quad \PB[P, Q]{\norm{\hat{\estParam}\left(Y_{1:n} \right)- \estParam \left( P \right)}_2^2 \leq \minimaxThres} \geq 1 - \failureProb \right\}.
    \end{align}
    \else
    \begin{equation}
        \minimaxQ{\failureProb, \PP, \eiVec} \coloneqq \underset{Q \in \QQ_{\eiVec}}{\inf} \; \underset{\hat{\estParam} \in \hat{\estParamClass}}{\inf} \; \underset{P \in \PP}{\sup} \; \inf \left\{ \minimaxThres \in [0, \infty) \; : \; \PB[P, Q]{\norm{\hat{\estParam}\left(Z_{1:n} \right)- \estParam \left( P \right)}_2^2 \leq \minimaxThres} \geq 1 - \failureProb \right\}.
    \end{equation}
    \fi
\end{definition}

By lower bounding the minimax quantile for a given problem by $\minimaxThres$, we are saying that no algorithm can achieve error better than $\minimaxThres$ with probability at least $1-\failureProb$ on that problem and over all distributions in family $\PP$.

\cite{Ma2024} also introduce methods that may be used to lower bound minimax quantiles. Specifically, they introduce a high-probability analogue of Le Cam's Lemma and give results that allow us to translate from expectation lower bounds to lower bounds on minimax quantiles. Our results use these techniques, among others.


\subsection{Notation}
We denote by $[n]$ the set $\{1, \ldots, n\}$.
We denote by $\mathbb{S}^{d-1}(r)$ and $\mathbb{B}^{d}(r)$  the Euclidean sphere and ball respectively of radius $r$ in to $\reals^d$.
We use the notation $\estParam\left( P \right)$ to indicate that the distribution $P$ is parametrized by $\estParam$. In the context of our work, this means that $\estParam$ is the expected value of $P$.

\subsection{Related Works}
Differential privacy, introduced by \citep{Dwork2006}, has been extensively studied across both the central and local model \citep{hsu2014, Bassily2015, Bassily2017, acharya2019hadamard, asoodeh2021local, canonne2024}. We refer the reader to detailed surveys for more information on the breadth of applications \citep{cormode2018survey, wang2020survey, yang2020survey}. 

We are primarily interested in the heterogeneous setting where users have an individual privacy preference. \cite{alaggan2015} and \cite{jorgensen2015} independently introduced this notion of heterogeneous differential privacy, also known as Personalized Differential Privacy in the central model. \cite{ChenLKJ16} first extended this notion to the local model. \cite{fallah2022bridging} studied the private mean estimation problem with heterogeneous privacy demands under the R\'enyi DP setting in both the local and central models, with results holding in expectation.
\cite{canonne2023} studied the heterogeneous setting in the local and shuffle models, although in their work there are only two possible privacy parameters. \cite{chaudhuri2025} also studied the heterogeneous setting, specifically the case in which a user's data is correlated with their privacy parameter. Our work examines the setting in which a user's privacy parameter is independent from their data.

Closely related to our work, we examine the application of LDP for mean estimation. 
\cite{ChenLKJ16} provided the first heterogeneous LDP algorithm for count estimation over a spatial domain. Our work differs from this result both in terms of setting and algorithm. While their aggregation technique is tailored to the hierarchical structure of spatial data, we develop a protocol well-suited to data from an arbitrary distribution. \cite{fallah2024optimal} presents optimal mean estimators for heterogeneous privacy loss levels with matching minimax lower bounds in both the central and local model. Their results apply to the single-dimensional setting and hold in expectation. In contrast, we propose high probability bounds for both the single-dimensional and multi-dimensional settings. \cite{chaudhuri2023free}, \cite{chaudhuri2023demands} presents the ``saturation phenomenon'' which demonstrates that relaxing the privacy requirements for some users while keeping the privacy parameter of the others fixed beyond a critical point does not improve accuracy of a heterogeneous mean estimator. 

Further, we also study the problem of distribution learning. While we focus on estimating the true proportion of an underlying population, this problem is similar to frequency estimation and the identification of ``heavy hitters''. Literature in this area focuses on optimizing various parameters including the worst-case error, time complexity for each user, time complexity at the server, and communication complexity. \cite{Bassily2015} presented the first LDP heavy hitters algorithm with optimal worst-case error. \cite{Bassily2017} then proposed an algorithm which significantly improved both server and user runtime. Both of these results rely on shared randomness which is referred to as the ``public-coin'' setting. \cite{acharya2019hadamard} developed a ``private-coin'' (i.e., no reliance on shared randomness) construction that achieved sample order optimality, logarithmic communication, and nearly linear runtime. Recently, \cite{canonne2024} examined the medium- or low-privacy (large $\varepsilon$) regime departing from the commonly studied high-privacy (small $\varepsilon$) regime and presented near-tight bounds on error for distribution learning.


\section{Single-Dimensional Mean Estimation}\label{sec:one-dim}

In this section, we study the heterogeneous LDP mean estimation problem for bounded distributions of a single dimension.
We assume that each of $n$ users holds a data point $X_i$ drawn from a distribution $P$ over a bounded domain $[-1,1]$. Each user passes their data through an $\ei$-LDP channel before sending it to the server. The server then estimates the mean of $P$, $\theta$, by aggregating each user's data. 
We find that different choices of $\ei$ across users would result in suboptimal tail bounds on the error if the server were to output an unweighted mean of the privatized data points. To negate this effect, we propose that the server computes an weighted average of the users' data, with weights chosen according to each user's privacy parameter.

We give matching upper and lower bounds for the heterogeneous-LDP mean estimation problem. We provide two algorithms achieving our upper bounds: the first, Algorithm~\ref{alg:lap}, applying to all distributions over $[-1, 1]$; and the second, Algorithm~\ref{alg:RR}, which specifically applies to binary distributions and achieves a low communication cost. We construct a hard instance and apply a high-probability instance of Le Cam's Lemma \citep{Ma2024} to prove our lower bound.

\subsection{Upper Bounds for Bounded Distributions}

Algorithm~\ref{alg:lap} uses a Laplace mechanism to guarantee the privacy of each user and outputs a weighted sum of each user's privatized data. Using standard concentration bounds, we give tail bounds of its error.

\begin{algorithm}[t]
\caption{Laplace Mechanism for Heterogeneous Local Differential Privacy}
\label{alg:lap}
\begin{algorithmic}[1]
    \Require Each user $i$ has private data $X_i \in [-1,1]$ and privacy parameter $\ei > 0$.
    \Ensure Estimate $\hat{\theta}$.
    \For{each user $i = 1, \dots, n$}
        \State Draw $Z_i \sim \Lap(2/\ei)$.
        \State $Y_i \gets X_i + Z_i$.
        \State Send $Y_i$ to the server.
    \EndFor
    \State Server sets $w_i \gets \left(1+\frac{1}{\ei^2}\right)^{-1} / \sum_{j=1}^n \left(1+\frac{1}{\epsDP_j^2}\right)^{-1}$ for all $i$.
    \State Server outputs $\hat{\theta} \gets \sum_{i=1}^n w_i Y_i$.
\end{algorithmic}
\end{algorithm}

\begin{restatable}{thm}{laplaceMech}
\label{thm:laplace_mech}
Let $\eiVec = \eis$, with $\ei \leq 1$ for all $i$. Let $\failureProb \in (0, 1)$. Let $\PP$ be the family of distributions $P$ such that for any $X \sim P$, $X \in [-1, 1]$ almost surely. There exists an $\eiVec$-locally differentially private estimator $\hat{\estParam}$ and a universal constant $c$ such that, for all $P \in \PP$,
\ifaistats
\begin{align}
    &\mathrm{\mathbf{Pr}}_{X_{1:n} \sim P^{\otimes n}}\left[ \abs{ \hat{\estParam}\left( X_{1:n} \right) - \estParam \left( P \right) }^2 \right. \nonumber \\
    & \left. \quaaad\quaaad \leq \min \left( c \;  \frac{\log \left( 1 / \failureProb \right)}{\sum_{i=1}^n \ei^2} , 1 \right) \right] \geq 1 - \failureProb.  
\end{align}
\else
\begin{equation}
    \PB[X_{1:n} \sim P^{\otimes n}]{\abs{ \hat{\estParam}\left( X_{1:n} \right) - \estParam \left( P \right) }^2 
    \leq \min \left( c \;  \frac{\log \left( 1 / \failureProb \right)}{\sum_{i=1}^n \ei^2} , 1 \right) } \geq 1 - \failureProb.  
\end{equation}
\fi

Algorithm~\ref{alg:lap} achieves this bound.
\end{restatable}


\subsection{Upper Bounds for Discrete Distributions}

The guarantees of Algorithm~\ref{alg:lap} apply to any bounded single-dimensional distribution, including binary distributions. 
However, this mechanism requires that each user communicates a continuous value, $Y_i$, with the server, regardless of whether that user's data was originally continuous. 
When each user holds only one bit of data, it is desirable to design a more communication-efficient mechanism, wherein each user only communicates one bit with the server.

To address this challenge, we propose a heterogeneous LDP mean estimation mechanism, based on the randomized response mechanism.
Specifically, assume each $X_i \sim P$, where $\E{P} = \theta \in [-1, 1]$ and $X_i \in \{-1, +1\}$. 
We construct a mean estimation mechanism that requires only one bit of communication per user, while achieving the same high-probability error guarantees as Algorithm~\ref{alg:lap}.

As in the previous algorithm, the proposed mechanism involves user $i$ sending the server an $\ei$-private copy of their data. The server then outputs a weighted sum of their data, with weights determined according to each user's privacy level. This mechanism is presented in Algorithm~\ref{alg:RR}. In Theorem~\ref{thm:RR}, we establish tail bounds on its error.

\begin{algorithm}[t]
\caption{Randomized Response Mechanism for Heterogeneous Local Differential Privacy}
\label{alg:RR}
\begin{algorithmic}[1]
    \Require Each user $i$ has private data $X_i \in \{-1,1\}$ and privacy parameter $\ei > 0$.
    \Ensure Estimate $\hat{\theta}$
    \For{each user $i = 1, \dots, n$}
        \vspace{1ex}
        \State $Y_i \leftarrow
            \begin{cases}
                X_i & w.p. \quad \frac{e^{\ei}}{e^{\ei}+1}\\
                -X_i & w.p.  \quad \frac{1}{e^{\ei}+1}\\
            \end{cases}. $
        \vspace{1ex}
        \State Send $Y_i$ to the server.
    \EndFor
    \State Server sets $c_i \leftarrow \frac{e^{\ei} + 1}{e^{\ei} - 1}$ for all $i$.
    \State Server sets $w_i \gets (1/c_i^2) / \sum_{j=1}^n (1/c_j^2)$ for all $i$.
    \State Server outputs $\hat{\theta} \gets \sum_{i=1}^n w_i c_i Y_i$.
\end{algorithmic}
\end{algorithm}

\begin{restatable}{thm}{RR}
\label{thm:RR}
Let $\eiVec = \eis$, with $\ei \leq 1$ for all $i$. Let $\failureProb \in (0, 1)$. Let $\PP$ be the family of distributions $P$ such that for any $X \sim P$, $X \in \{-1, 1\}$ almost surely. There exists an $\eiVec$-locally differentially private estimator $\hat{\estParam}$ and a universal constant $c$ such that, for all $P \in \PP$,
\ifaistats
\begin{align}
    &\mathrm{\mathbf{Pr}}_{X_{1:n} \sim P^{\otimes n}}\left[ \abs{ \hat{\estParam}\left( X_{1:n} \right) - \estParam \left( P \right) }^2 \right. \nonumber \\
    & \left. \quaaad\quaaad \leq \min \left( c \;  \frac{\log \left( 1 / \failureProb \right)}{\sum_{i=1}^n \ei^2} , 1 \right) \right] \geq 1 - \failureProb.  
\end{align}
\else
\begin{equation}
    \PB[X_{1:n} \sim P^{\otimes n}]{\abs{ \hat{\estParam}\left( X_{1:n} \right) - \estParam \left( P \right) }^2 \leq \min\left( c \; \frac{\log \left( 1 / \failureProb \right)}{\sum_{i=1}^n \ei^2} , 1 \right) } \geq 1 - \failureProb. 
\end{equation}
\fi

Algorithm~\ref{alg:RR} achieves this bound.
\end{restatable}


\subsection{Lower Bounds}
\label{sec:lower-bounds}
In this section, we establish a lower bound on the high probability error of one-dimensional mean estimation under heterogeneous LDP.
Up to a constant factor, the upper bounds presented in Theorem~\ref{thm:laplace_mech} and Theorem~\ref{thm:RR} for the heterogeneous Laplace mechanism and randomized response, respectively, match this lower bound, thereby characterizing the tightness of these results. 

Our proof constructs a hard instance consisting of a pair of distributions, $\muOne$ and $\muTwo$, whose means differ by a fixed gap. We apply a strong data processing inequality of \cite{duchi2013local} to show that passing these distributions through homogeneous privacy channels reduces their KL divergence, making them harder to distinguish. We then invoke a high probability version of Le Cam's Lemma, due to \cite{Ma2024}, which implies that, because the privatized distributions are nearly indistinguishable, no algorithm can estimate their means accurately with high probability.

\begin{restatable}{thm}{lowerBound}\label{thm:lowerBound}
Let $\eiVec = \eis$, with $\ei \leq 1$ for all $i$. Let $\failureProb \in (0, 1/2)$. Let $\PP$ be the family of distributions $P$ such that for any $X \sim P$, $X \in \{-1, 1\}$ almost surely. For the $\eiVec$-locally differentially private mean estimation problem over a single dimension, there exists an absolute constant $c$ such that the minimax quantile is lower bounded as
     \begin{equation} \label{eqn:lower_bound2}
         \min \left( c \; \frac{\log(1 / \failureProb)}{\sum_{i=1}^n \ei^2}, 1\right) \leq \minimaxQ{\failureProb, \PP, \eiVec}.
     \end{equation}
where $c>0$ is a universal constant.
\end{restatable}

\section{Multi-Dimensional Mean Estimation} \label{sec:high-dim}
In this section, we study the heterogeneous LDP mean estimation problem for distributions over the Euclidean ball of radius $\radius$. We assume that each of $n$ users holds a datapoint $X_i \in \mathbb{B}^{\dimValues}\left(\radius \right)$, drawn from a distribution $P$ with mean $\estParam$. Each user passes their data through a locally private channel before sharing it with the untrusted server for aggregation. In Theorem~\ref{thm:highDimUB}, we give an upper bound for this problem, achieved by Algorithm~\ref{alg:highDim}. In Theorem~\ref{thm:highDimLB}, we prove a matching lower bound. 

\subsection{Upper Bounds}
Within Algorithm~\ref{alg:highDim}, each user employs a locally differentially private mechanism proposed by \cite{duchi2013local} (Algorithm~\ref{alg:duchi}). On input $X_i \in  \mathbb{B}^{\dimValues}\left(\radius \right)$, and given privacy parameter $\ei$, this mechanism outputs $\userOut \in \mathbb{S}^{\dimValues-1}\left(\boundi \right)$ for some $\boundi = \mathcal{O}(\radius \dimValues / \ei)$. In particular, the mechanism picks either the hemisphere $\left \{Y \in \mathbb{S}^{\dimValues - 1}(\boundi) \; \text{ s.t. } \left\langle Y, X_i \right\rangle > 0\right\}$ with some probability $p_i \geq 1/2$, or its complementary hemisphere with the complementary probability, and then samples uniformly from the chosen hemisphere. The choice of $p_i$ ensures privacy, and together with the choice of $\boundi$ makes the output unbiased, i.e., $\E{\userOut} = X_i$.

Lastly, as in the single-dimensional setting, the server outputs a weighted average of the users' projections, with the weight of user $i$ being proportional to $\ei^2$.

\begin{algorithm}[t]
\caption{Optimal Multi-Dimensional Mean Estimation Mechanism under Heterogeneous Local Differential Privacy}
\label{alg:highDim}
\begin{algorithmic}[1]
    \Require Each user $i$ has private data $X_i \in \mathbb{B}^{\dimValues}\left(\radius \right)$, $X_i \sim P$, and privacy parameter $\ei > 0$.
    \Ensure Estimate $\hat{\theta}$ of $\E{P}$.
    \For{each user $i = 1, \dots, n$}
        \State $\userOut \leftarrow$ output of Strategy~A of \cite{duchi2013local} (Algorithm~\ref{alg:duchi}) on $X_i$, $\ei$.
        \State Send $\userOut$ to the server.
    \EndFor
    \State Server sets $w_i \gets \ei^2 / \sum_{j=1}^n \epsDP_j^2$ for all $i$.
    \State Server outputs $\hat{\theta} \gets \sum_{i=1}^n w_i \userOut$.
\end{algorithmic}
\end{algorithm}

\begin{algorithm}[t]
\caption{Locally-Differentially Private Randomizer of \cite{duchi2013local}}
\label{alg:duchi}
\begin{algorithmic}[1]
    \Require Data point $X_i \in \mathbb{B}^{\dimValues}\left(\radius \right)$, privacy parameter $\ei > 0$.
    \Ensure $\userOut$, an $\ei$-locally differentially private estimate of $X_i$.
    \vspace{1ex}
    \State $\Tilde{X}_i \leftarrow
        \begin{cases}
            \frac{\radius  X_i}{\norm{X_i}_2} & w.p. \; \frac{1}{2} + \frac{\norm{X_i}_2}{2 \radius} \\
            -\frac{\radius  X_i}{\norm{X_i}_2} & w.p. \; \frac{1}{2} - \frac{\norm{X_i}_2}{2 \radius}\\
        \end{cases}. \label{line:highDimAlgChoiceXTilde}$
    \vspace{1ex}
    \State $T \sim \Bern\left(\frac{e^{\ei}}{e^{\ei}+1}\right)$.
    \vspace{1ex}
    \State $c_i \leftarrow \frac{e^{\ei}+1}{e^{\ei}-1}$.
    \vspace{1ex}
    \State $\boundi \leftarrow c_i r \frac{\dimValues \sqrt{\pi} \Gamma\left( \frac{\dimValues-1}{2} + 1 \right)}{\Gamma\left( \frac{\dimValues}{2} + 1 \right)}$.
    \vspace{1ex}
    \If{$T = 1$}
    \vspace{0.2ex}
    \State $\userOut \sim 
            \Unif \left( Y \in \mathbb{S}^{\dimValues - 1}(\boundi) \; \text{ s.t. } \left\langle Y, \Tilde{X}_i \right\rangle > 0 \right).$
    \vspace{0.1ex}
    \ElsIf{$T=0$}
    \vspace{0.2ex}
    \State $\userOut \sim \Unif \left( Y \in \mathbb{S}^{\dimValues - 1}(\boundi) \; \text{ s.t. } \left\langle Y, \Tilde{X}_i \right\rangle \leq 0\right)$. \label{line:highDimAlgChoiceY}
    \EndIf
    \State Output $\userOut$.
\end{algorithmic}
\end{algorithm}



\begin{restatable}{thm}{highDimUB}
\label{thm:highDimUB}
Let $\eiVec = \eis$, with $\ei \leq 1$ for all $i$. Let $\failureProb \in (0, 1)$. Let $\PP_{2, \radius}$ be the family of distributions $P$ such that for any $X \sim P$, $X \in \mathbb{B}^{\dimValues}\left(\radius \right)$ almost surely. There exists an $\eiVec$-locally differentially private estimator $\hat{\estParam}$ and a universal constant $c$ such that, for all $P \in \PP_{2, \radius}$,
\ifaistats
\begin{align}
    &\mathrm{\mathbf{Pr}}_{X_{1:n} \sim P^{\otimes n}}\left[ \norm{ \hat{\estParam}\left( X_{1:n} \right) - \estParam \left( P \right) }_2^2 \right. \nonumber \\
    & \left. \quad \leq r^2 \min \left( c \frac{ \left( \dimValues + \log \left( 1 / \failureProb \right)\right)}{\sum_{i=1}^n \ei^2} , 1 \right) \right] \geq 1 - \failureProb.  
\end{align}
\else
\begin{equation}
    \PB[X_{1:n} \sim P^{\otimes n}]{\norm{ \hat{\estParam}\left( X_{1:n} \right) - \estParam \left( P \right) }_2^2 \leq r^2 \min \left( c \frac{ \left( \dimValues + \log \left( 1 / \failureProb \right)\right)}{\sum_{i=1}^n \ei^2} , 1 \right)} \geq 1 - \failureProb. 
\end{equation}
\fi
Algorithm~\ref{alg:highDim} achieves this bound.
\end{restatable}




\begin{proof}[Proof Sketch]
A natural attempt, similar to what we did in the single-dimensional case, would be to apply concentration inequalities for high-dimensional random variables. Given the boundedness of the output of the privacy channel, i.e., $Y_i$, two natural candidates are the Vector Bernstein inequality \citep{gross2011recovering} and subgaussian bounds for bounded-support distributions. One can verify that both approaches lead to an upper bound with dependence on $\dimValues \log \left( 1/\failureProb \right)$ (rather than what we ultimately show, which is $\dimValues + \log \left( 1/\failureProb \right)$), which is suboptimal in the high-probability regime. 

We resolve this issue by taking advantage of the structure of the $\ei$-LDP randomization strategy of \cite{duchi2013local} (Algorithm~\ref{alg:duchi})---specifically, that it draws $\userOut$ from a mixture of uniform distributions over hemispheres. As we will discuss, uniform distributions over spheres exhibit the \textit{concentration of measure} phenomenon, which allows us to establish tighter concentration inequalities.

Our point of departure in the proof is to break the problem of deriving high-probability bounds into two separate tail bounds. The first concerns the sampling (non-private) error of the algorithm, i.e., the difference between the weighted average of the $X_i$’s and $\estParam$. The second concerns the error due to privacy, i.e., the difference between the the weighted average of the $X_i$’s and the weighted average of the $Y_i$’s, for any realization of the $X_i$’s.

To bound the sampling error, we use the \textit{norm-subgaussian} definition from \cite{jin2019} (also stated in the appendix for completeness). In particular, the norm of each vector $X_i - \estParam$ is bounded by $2r$, which implies that $X_i - \estParam$ is a norm-subgaussian random vector with parameter $2r$. We can then apply the Hoeffding-like concentration inequality developed by \cite{jin2019}, which allows us to establish a bound of $\radius^2 \log(\dimValues / \failureProb)/{\sum_i \ei^2}$.

Bounding the tail of the privacy error (conditioned on realization of $X_i$'s) is more nuanced. In particular, we show that, for any fixed $i$, $\userOut - X_i$ is subgaussian, with a subgaussian norm \textit{independent of $\dimValues$}, which in turn leads to the desired high-probability bounds by applying norm concentration results for subgaussian vectors (e.g., see \citep{wainwright2019high, liu2025new}).
Recall that a random vector $Z$ is subgaussian if $\langle Z, \ell \rangle$ is subgaussian for all $\ell$, and the subgaussian norm of $Z$ is defined as the supremum of the subgaussian norms of $\langle Z, \ell \rangle$ over all $\ell$.

We establish this result using Lévy’s lemma, a concentration-of-measure result on spheres that allows us to prove the sub-Gaussianity of any Lipschitz function of a uniformly distributed random vector on the sphere.
    
\begin{restatable}[Levy's Lemma {\citep[Theorem 5.1.3]{Vershynin2018}}]{lem}{levy}
\label{thm:Levy}
Let $\Unif$ be the uniform distribution over $\mathbb{S}^{\dimValues-1}(\bound)$.
Let $f: \mathbb{S}^{\dimValues-1}(\bound) \rightarrow \reals$ be an $\eta$-Lipschitz function. For all $t \geq 0$, there exists a universal constant $c$ such that
\begin{equation*}
\PB[Y \sim \Unif]{\abs{f\left(Y \right) - \E{f\left(Y\right)}} \geq t} \leq \exp \left( - c \frac{\dimValues t^2}{B^2 \eta^2}\right).
\end{equation*}
\end{restatable}
    
This lemma allows us to establish subgaussian tail bounds on $f(Y) - \E{f(Y)}$ with $f(Y) = \langle \ell, Y\rangle$ when $Y$ is drawn uniformly over $S = \mathbb{S}^{\dimValues-1}(\boundi)$. However, in our setting, $\userOut$ is not uniformly drawn over $S$. That said, we can interpret the distribution of $\userOut$ as a mixture of a uniform distribution over $S$ and a uniform distribution over $S_1$, where $S_1$ is the half-sphere of $S$ defined by $\langle Y, X_i \rangle \geq 0$. 
Our proof proceeds in two steps: first, we show that $f(Y) - \E{f(Y)}$ is subgaussian even when $Y$ is drawn uniformly from the hemisphere $S_1$; second, combining the two results, we establish that the distribution of $\langle \ell, \userOut - X_i \rangle$ is subgaussian when $\userOut$ is the output of the privacy channel of \Cref{alg:duchi}.

To prove the first step, we define a function $g(Y)$ on the entire sphere $S$ as follows: on $S_1$, let $g(Y) = \langle \ell, Y \rangle$; on $S_2$, first reflect $Y$ into $S_1$ with respect to the plane separating $S_1$ and $S_2$ to obtain $Y'$, and then let $g(Y) = \langle \ell, Y' \rangle$. Applying Lemma~\ref{thm:Levy} to the function $g(\cdot)$ implies that the centered version of $\langle \ell, Y \rangle$ is subgaussian when $Y$ is drawn uniformly from the hemisphere $S_1$.


For the second step, while we have shown that $f(Y) - \E{f(Y)}$ is subgaussian when $Y$ is drawn uniformly from $S$ or from $S_1$, the mean $\E{f(Y)}$ differs between these cases. This difference prevents us from immediately combining the tail bounds of both distributions to establish a tail bound on $f(Y) - \E{f(Y)}$ when $Y = \userOut$.
Instead, we use the definition of subgaussianity based on moment-generating functions and show that the error introduced by these different means can be controlled by a constant factor. This, in turn, establishes the subgaussianity of $f(\userOut) - \E{f(\userOut)}$ and completes the proof.
\end{proof}

\subsection{Lower Bounds}
We give a lower bound on the high probability error of multi-dimensional mean estimation under heterogeneous LDP.
This lower bound matches the upper bound presented in Theorem~\ref{thm:highDimUB}, demonstrating the optimality of Algorithm~\ref{alg:highDim}.

The proof of this bound involves constructing a hard instance, consisting of distributions parametrized by the vertices of the $\dimValues$-dimensional hypercube. We show that the distributions corresponding to any two adjacent vertices become nearly indistinguishable after they are passed through privacy channels. This allows us to apply Assouad's Lemma, therefore lower bounding the minimax risk of mean estimation over this class of distributions. To connect this bound on minimax risk to minimax quantiles, we apply Theorem~\ref{thm:expToHighProb} \citep{Ma2024}. 

\begin{restatable}{thm}{highDimLB}
\label{thm:highDimLB}
Let $\eiVec = \eis$, with $\ei \leq 1$ for all $i$. Let $\failureProb \in (0, 3/20)$. Let $r > 0$. For all $i \in [n]$, let $\ei \in (0, 1)$. For the mean estimation problem over the $\ell_2$ ball with radius $\radius$, there exists an absolute constant $c$ such that the minimax quantile is lower bounded as
\ifaistats
    \begin{align*}
        &\minimaxQ{\failureProb, \PP_{2, \radius}, \eiVec} \geq \\
        & c ~r^2 \min \left( 
        \frac{ \log ({1}/{\failureProb})+d}{\sum_{i=1}^n \ei^2},
        \frac{ \log ({1}/{\failureProb})}{\sum_{i=1}^n \ei^2} + \frac{ 1}{\sqrt{\sum_{i=1}^n \ei^2}},
        1  \right).
    \end{align*}
\else
    \begin{equation*}
        \minimaxQ{\failureProb, \PP_{2, \radius}, \eiVec} \geq
         c ~r^2 \min \left( 
        \frac{ \log ({1}/{\failureProb})+d}{\sum_{i=1}^n \ei^2},
        \frac{ \log ({1}/{\failureProb})}{\sum_{i=1}^n \ei^2} + \frac{ 1}{\sqrt{\sum_{i=1}^n \ei^2}},
        1  \right).
    \end{equation*}
\fi
\end{restatable}

\section{Distribution Learning} \label{sec:hist-est}



In this section, we demonstrate an adaptation of the projection-based homogeneous locally private frequency estimation algorithm of \cite{Bassily2015} for heterogeneous locally private distribution learning. Their algorithm produces a computation- and communication-efficient frequency oracle by employing a Johnson-Lindenstrauss (JL) transform to project the histogram of observed values to a lower-dimensional space. 

More precisely, each user compresses their data point via a random
$m \times d$ matrix $\Phi \sim \Unif\left\{ - 1/{\sqrt{m}}, 1/{\sqrt{m}} \right\}$. Each user then selects one bit of their compressed vector, runs randomized response to privatize this bit, and shares the resulting bit with the server, which aggregates all bits. 
The Johnson-Lindenstrauss Lemma (Lemma~\ref{lem:JL}) ensures that the error due to this projection is not too large.

\begin{restatable}[Johnson-Lindenstrauss Lemma]{lem}{JL}
    \label{lem:JL}
    Let $0 < \lambda < 1$ and $d \in \mathbb{N}$. For any set $\VV$ of $t$ points in $\mathbb{R}^d$ and $m \geq \frac{8 \log \p{t}}{\lambda ^2}$, there exists a linear map $\Phi: \mathcal{R}^d \rightarrow \mathcal{R}^m$ such that for all $\mathbf{x}, \mathbf{y} \in \VV$, we have 
     \[
        (1 - \lambda) \norm{ \mathbf{x} - \mathbf{y} }_2^2 \leq \norm{\Phi (\mathbf{x} - \mathbf{y})}_2^2 \leq (1 + \lambda) \norm{ \mathbf{x} - \mathbf{y}}_2^2
    \]
    and
     \[
        \abs{ \dotp{ \Phi \mathbf{x}, \Phi \mathbf{y} } - \dotp{ \mathbf{x}, \mathbf{y}} } \leq O\p{ \lambda \p{ \norm{\mathbf{x}}_2^2 + \norm{\mathbf{y}}_2^2}}.
    \]
\end{restatable}

Since we work in the heterogeneous LDP setting, we must adjust the server’s aggregation step compared to the homogeneous case. We present the modified algorithm in Algorithm~\ref{alg:histEstBassily}. In Theorem~\ref{thm:HistEstBassilyUB}, we give guarantees on its $\ell_\infty$-error for distribution learning.

\begin{algorithm}[t]
\caption{Distribution Learning Mechanism Under Local Heterogeneous Differential Privacy}
\label{alg:histEstBassily}
\begin{algorithmic}[1]
    \Require Each user $i$ has private data $\obsValue_i \in 
    [\dimValues]$ and privacy parameter $\ei > 0$. Server has a global confidence parameter $\failureProb \in (0, 1)$.
    \Ensure Estimates $\hat{p}(\trueValue)$ for all $\trueValue \in [\dimValues]$.
    \vspace{1ex}
    \State $c_i \leftarrow \frac{e^{\ei} + 1}{e^{\ei} - 1}$ for all $i$.
    \vspace{1ex}
    \State $\JLconstant \leftarrow  \frac{ \sqrt{\log \left( {2\dimValues} / {\failureProb} \right) }}{  \sum_{i=1}^n {1} / {c_i^2} }$.
    \vspace{1ex}
    \State $\dimCompressed \leftarrow \frac{\log \left( \dimValues + 1 \right) \log \left(2 / \failureProb \right)}{\JLconstant^2}$.
    \State Server generates $\projMatrix \sim \Unif\left( \{ -\frac{1}{\sqrt{\dimCompressed}} , \frac{1}{ \sqrt{\dimCompressed}} \}^{\dimCompressed \times \dimValues}\right)$. 
    \For{each user $i = 1, \dots, n$}
        \State $\privateCompVecI \leftarrow \textsc{local-randomizer}\left( x_i, \ei, \projMatrix , c_i \right)$
        \State Send $\privateCompVecI$ to the server.
    \EndFor
    \State Server sets $w_i \gets (1/c_i^2) / \sum_{j=1}^n (1/c_j^2)$ for all $i$.
    \State Server computes $\compressedMean \gets \sum_{i=1}^n w_i \; \privateCompVecI$. \label{line:compressedMean}
    \State Server outputs $\hat{p}(\trueValue) = \left\langle \compressedMean, \projMatrix \basisVec_\trueValue \right\rangle$ for all $\trueValue \in [\dimValues]$. \label{line:histEstFinalOut} 
    \vspace{2ex}
    \Statex {The following subroutine is due to \cite{Bassily2015}.}
    \Procedure{Local-Randomizer}{$x_i$, $\ei$, $\projMatrix$, $c_i$}
        \State $\compressedVecI \leftarrow \projMatrix \basisVec_{\obsValue_i}$.
        \State Sample $j \sim \Unif\left( [\dimCompressed] \right)$.
        \If{$ \compressedVecI \ne \mathbf{0} $}
            \vspace{1ex}
            \State $\privateCompVecI{}_j \leftarrow             \begin{cases}
                c_i \dimCompressed \; \compressedVecI{}_j & w.p. \quad \frac{e^{\ei}}{e^{\ei}+1}\\
                - c_i \dimCompressed \; \compressedVecI{}_j & w.p.  \quad \frac{1}{e^{\ei}+1}\\
            \end{cases}. $
            \vspace{1ex}
        \Else
            \State $\privateCompVecI{}_j \sim \Unif \left( \left\{ -c_i \sqrt{\dimCompressed}, c_i \sqrt{\dimCompressed} \right\} \right)$.
        \EndIf
        \State Set $\privateCompVecI{}_\ell = 0$ for all $\ell \ne j$. 
        \State Send $\privateCompVecI$ to the server.
    \EndProcedure
\end{algorithmic}
\end{algorithm}




\begin{restatable}{thm}{HistEstBassilyUB}
\label{thm:HistEstBassilyUB}
    Let $\eiVec = \eis$, with $\ei \leq 1$ for all $i$. Let $\failureProb \in (0, 1)$. Let $\PP$ be the family of distributions $P$ over $[d]$. There exists an $\eiVec$-locally differentially private estimator $\hat{P}$ and a universal constant $c$ such that, for all $P \in \PP$,
    \ifaistats
    \begin{align}
        &\mathrm{\mathbf{Pr}}_{\obsValue_{1:n} \sim P^{\otimes n}}\left[ \norm{\hat{P}\left( \obsValue_{1:n} \right) - P}_\infty \right. \nonumber \\
        & \left. \quaaad \leq \min \left( c \; \sqrt{ \frac{\log \left(\dimValues / \failureProb \right)}{\sum_{i=1}^n \ei^2}} , 1 \right) \right] \geq 1 - \failureProb.  
    \end{align}
    \else
    \begin{equation}
        \PB[\obsValue_{1:n} \sim P^{\otimes n}]{\norm{\hat{P}\left( \obsValue_{1:n} \right) - P}_\infty \leq \min \left( c \; \sqrt{ \frac{\log \left(\dimValues / \failureProb \right)}{\sum_{i=1}^n \ei^2}} , 1 \right)} \geq 1 - \failureProb.
    \end{equation}
    \fi
    Algorithm~\ref{alg:histEstBassily} achieves this bound.
\end{restatable}

We note that Algorithm~\ref{alg:histEstBassily} is not necessarily a proper learning algorithm, as there is no guarantee that the output $\hat{p}$ is a valid probability distribution. However, in Theorem~\ref{thm:HistEstBassilyUB}, we show that it closely approximates the true distribution $p$ in $\ell_\infty$ distance. To transform $\hat{p}$ into a probability distribution, one could project it onto the probability simplex without increasing the algorithm's error or sacrificing privacy.

Further, we remark that Algorithm~\ref{alg:histEstBassily} can be employed either to provide an estimate of the density of the entirety $p$ or as an oracle for estimating the density of any given value, $\trueValue$, on demand. To perform the latter task, the algorithm does not need to execute Line~\ref{line:histEstFinalOut} for all $\trueValue \in [d]$, therefore improving its computational efficiency.

\section{Conclusion}
\label{sec:conclusion}
We address key statistical estimation problems under local differential privacy in the heterogeneous setting, where users may have different privacy budgets and where error guarantees must hold with high probability. For bounded single-dimensional mean estimation, we show that the optimal high-probability error is achieved by combining a privacy-weighted aggregation technique with the Laplace mechanism or the randomized response mechanism. In the multi-dimensional setting, for distributions with bounded norm, we demonstrate that the locally private mechanism of \cite{duchi2013local}, when paired with a carefully-chosen weighted aggregation technique, is optimal in the heterogeneous high-probability regime.

These results highlight that, with only slight modifications, well-understood mechanisms can serve as effective building blocks for more general heterogeneous LDP algorithms. The high-probability guarantees we derive ensure that these mechanisms can be composed across multiple invocations, as is often required in more complex settings. We illustrate one such application in distribution estimation. Future work may explore applying these techniques to other statistical estimation tasks, such as frequency estimation.







\bibliography{References}
\newpage


\appendix

\section{Background Results} 
In the following, we list useful facts, definitions, lemmas and theorems used throughout our proofs.

\subsection{Subgaussian and Subexponential Random Variables}
We give the definitions and key properties of subexponential and subgaussian random variables, as well as norm-subgaussian random vectors, used throughout our work. For further background on subexponential and subgaussian random vectors, see \citep{Vershynin2018}. For further background on norm-subgaussian random vectors, see \cite{jin2019}.

\vspace{1em}

\begin{definition}[Subexponential random  variable] \citep{Vershynin2018}
\label{def:subexp}
Let $\nu, \alpha > 0$. A random variable with mean $\E{X} = \mu$ is $(\nu^2, \alpha)$-subexponential, denoted $X \in subE(\nu^2, \alpha)$ if
\[ \EX{e^{\lambda(X - \mu)}} \leq e^{\frac{\nu^2 \lambda^2}{2}} \tag{$\forall \lambda : \abs{\lambda} \leq \frac1{\alpha}$} \]
\end{definition}

\vspace{1em}

\begin{fact} \label{fct:subE_bound}
    If $X \in subE(\nu^2, \alpha)$ then, for all $t > 0$,
    \[ \PB{\abs{X - \EX{X}} \geq t} \leq \exp \p{- \frac1{2} \min \left\{ \frac{t^2}{\nu^2}, \frac{t}{\alpha} \right\}}.\]
\end{fact}


\vspace{1em}

\begin{definition}[Subgaussian random variable]
\label{def:subgaussian}
\citep{Vershynin2018} Let $\nu_1 > 0$. A random variable $X$ is said to be $\nu_1$-subgaussian if it satisfies:
\begin{equation} \label{eqn:subGaussianTail}
    \PB{\abs{X} \geq t} \leq 2 \exp \p{ \frac{-t^2}{\nu_1^2}} \, \forall t \geq 0.
\end{equation} 
If $\EX{X} = 0$, the property given in \eqref{eqn:subGaussianTail} is equivalent to the following for $\nu_2 = O(\nu_1)$:
\begin{equation}
\label{eqn:subgaussianMGF}
    \EX{\exp \p{\lambda X}} \leq \exp \p{\nu_2^2 \lambda^2} \, \forall \lambda \in \mathbb{R}.
\end{equation}

We say that $\nu_1$ (or $\nu_2$) is the \emph{variance proxy} of $X$.
We note that in \eqref{eqn:subGaussianTail}, the constant $2$ can be any absolute constant greater than $1$. Similarly, in \eqref{eqn:subgaussianMGF}, a multiplicative constant other than 1 may be had. Changing these constants changes the variance proxy by a constant factor. (Remark 2.6.3 of \cite{Vershynin2018}).
\end{definition}

\vspace{1em}

\begin{definition}[Subgaussian random vector]
\label{def:subgaussianvector}
\citep{Vershynin2018} Let $\nu > 0$. A random vector $X \in \reals^\dimValues$ is said to be $\nu$-subgaussian if $\langle \ell, X \rangle$ is $\nu$-subgaussian for all $\ell \in \mathbb{S}^{\dimValues - 1}$.
\end{definition}

\vspace{1em}

\begin{fact}[Norm of subgaussian random vectors] \citep{wainwright2019high, liu2025new} \label{fct:subGaussianNorm}
    For a subgaussian vector $X \in \reals^\dimValues$ with variance proxy $\sigma^2$, there exist constants $C_1, C_2$ such that for all $\failureProb \in (0, 1)$,
    \begin{equation}
        \PB{\norm{X}_2 \leq \sigma \sqrt{C_1 d + C_2 \log\left(1 / \failureProb \right)}} \geq 1 - \failureProb.
    \end{equation}
\end{fact}

\vspace{1em}

\begin{definition}
    \label{def:normSubGaussian}
    Let $\sigma > 0$. A random vector $X \in \reals^\dimValues$ is $\sigma$-norm-subgaussian if it satisfies, for all $t \in \reals$,
    \begin{equation}
        \PB{\norm{X - \E{X}}_2 \geq t} \leq 2 \exp \left( - \frac{t^2}{2\sigma^2}\right).
    \end{equation}
\end{definition} 

\vspace{1em}

\begin{fact}
    \label{fct:normSubGaussian}
    Let $\sigma > 0$. Let $X \in \reals^\dimValues$ be a random vector such that $\norm{X} \leq \sigma$. Then, $X$ is $\sigma$-norm-subgaussian.
\end{fact}

\vspace{1em}

\begin{fact}\citep{jin2019}
    \label{fct:normSubGHoeffdingBound}
    Let $\failureProb \in (0, 1)$. Let $X_1, \ldots, X_n \in \reals^\dimValues$ be random vectors. Assume $X_i$ is $\sigma_i$-norm-subgaussian for all $i \in [n]$. Then, there exists a constant $c$ such that:
    \begin{equation}
        \PB{\norm{\sum_{i=1}^n X_i}_2 \leq c \sqrt{\sum_{i=1}^n \sigma_i^2 \log \left( 2\dimValues / \failureProb \right)}} \geq 1 - \failureProb
    \end{equation}
\end{fact}

\subsection{Lower Bound Techniques}

\begin{lemma}[Assouad's Lemma \citep{yu1997assouad, Ma2024}]
    \label{lem:assouad}
    Let $k \geq 1$. Let $\VV = \{-1, 1\}^k$. For all $\nu, \nu' \in \VV$, write $\nu \sim \nu'$ if $\nu$ and $\nu'$ differ in only one coordinate and $\nu \sim_j \nu'$ when that coordinate is $j$. For $\nu \in \VV$, let $P_\nu \in \PP$. Suppose there are $k$ pseudo-distances $d_1, \ldots, d_k$ on $\estParamClass$ such that for any $\estParam_1, \estParam_2 \in \estParamClass$:
    \begin{equation}
        d \left(\estParam_1, \estParam_2 \right)
        = \sum_{j=1}^k d_j\left(\estParam_1, \estParam_2 \right).
    \end{equation}
    Then,
    \begin{align*}
        \inf_{\hat{\estParam} \in \hat{\estParamClass}} \;  \max_{\nu \in \VV} \; &\E[X_{1:n} \sim P_\nu^{\otimes n}]{d\left( \hat{\estParam}\left( X_{1:n} \right),  \estParam \left( P_\nu \right) \right)} \\
        &\geq \frac{k}{2} \cdot \min_{j \in [k], \nu \sim_j \nu'} d_j \left(\estParam(P_\nu), \estParam(P_{\nu'}) \right) \cdot \min_{\nu \sim \nu'} \left(1 - \TV{P_\nu}{P_{\nu'}} \right).
    \end{align*}
\end{lemma}

\vspace{1em}

\begin{theorem} [Pairwise upper bound on Kullback-Leibler divergences \citep{duchi2013local, Duchi2017}] \label{thrm:KL_bound} 
Let $\epsDP \geq 0$. Consider any pair of distributions $\mu^{(1)}$ and $\mu^{(2)}$. Let $P^{(1)}$ and $P^{(2)}$ be the pair of distributions induced by passing samples drawn from $\mu^{(1)}$ and $\mu^{(2)}$ through a $\epsDP$-LDP channel. Then we have
\begin{equation}
    \KL{P^{(1)}}{P^{(2)}} + \KL{P^{(2)}}{P^{(1)}} \leq \min \left\{ 4, e^{2\epsDP} \right\} \left(e^{\epsDP} - 1 \right)^2 \TVsquared{\muOne}{\muTwo}.
\end{equation}
\end{theorem}

\subsubsection{Minimax Quantiles}
\cite{Ma2024} introduce the concept of minimax quantiles (Definition~\ref{def:minimax_quantile}), which we use to construct our lower bounds. They further develop tools and techniques for working with minimax quantiles. Among these tools are lower minimax quantiles, defined in Definition~\ref{def:lowerMinimax}. Lower minimax quantiles are particularly useful as they lower bound minimax quantiles, as stated in Theorem~\ref{thm:expToHighProb}, and may be easier to work with. Corollary~\ref{cor:KL_bound} and Theorem~\ref{thm:expToHighProb} give examples of techniques which may be used to lower bound lower minimax quantiles. Corollary~\ref{cor:KL_bound} lower bounds this quantity whenever there exist two distributions in the family with sufficiently small KL divergence. Theorem~\ref{thm:expToHighProb} establishes a relationship between lower minimax quantiles and minimax risk.

\vspace{1em}

\begin{definition}[Lower minimax quantile \citep{Ma2024}] \label{def:lowerMinimax}
    Let $\eiVec = \eis$, with $\ei > 0$ for all $i$. Let $\failureProb \in (0, 1)$. Let $\PP$ be a family of distributions, where $P_\theta \in \PP$ is parametrized by $\theta \in \estParamClass$. 
    Let $\QQ_{\eiVec}$ be the set of all conditional distributions $Q: \XX^{\otimes n} \rightarrow \ZZ^{\otimes n}$ guaranteeing $\eiVec$-local differential privacy.
    Let $\hat{\estParamClass}$ be the set of all measurable functions $\hat{\estParam} : \ZZ^{\otimes n} \rightarrow \estParamClass$. 
    Then, we define the lower minimax quantile, $\lowerMinimaxQ{\failureProb, \PP, \eiVec}$ as:
    \begin{equation}
        \lowerMinimaxQ{\failureProb, \PP, \eiVec} \coloneqq \inf \left\{ \minimaxThres \in [0, \infty) \; : \; \underset{Q \in \QQ_{\eiVec}}{\inf} \; \underset{\hat{\estParam} \in \hat{\estParamClass}}{\inf} \; \underset{P \in \PP}{\sup} \; \PB[P, Q]{\norm{\hat{\estParam}\left(Z_{1:n} \right)- \estParam \left( P \right)}_2^2 \leq \minimaxThres} \geq 1 - \failureProb \right\}.
    \end{equation}
\end{definition}

\vspace{1em}

\begin{theorem}[Theorem 4 of \citep{Ma2024}, adapted to our notation]
\label{thrm:quantile_bound}
    For all $\failureProb \in (0, 1]$, families of distributions $\PP$ and privacy budgets $\eiVec = \eis$,
    \begin{equation}
        \lowerMinimaxQ{\failureProb, \PP, \eiVec} \leq \minimaxQ{\failureProb, \PP, \eiVec}.
    \end{equation}
\end{theorem}

\vspace{1em}

\begin{corollary}[Corollary 6 of \cite{Ma2024}, adapted to our setting] 
\label{cor:KL_bound}
Let $\eiVec = \eis$, with $\ei \in (0, 1]$ for all $i$. Let $\failureProb \in (0, 1/2)$. Let $\bm{\mathcal{C}}$ be an $\eiVec$-privacy channel. Suppose that $\muOne \in \PP$ and $\muTwo \in \PP$ satisfy $\KL{\bm{\mathcal{C}}\left(\muOne\right)}{\bm{\mathcal{C}}\left(\muTwo\right)} <  \log{ ( 1 / (4 \failureProb (1-\failureProb)))}$. Then, 
\begin{equation}
    \lowerMinimaxQ{\failureProb, \PP, \eiVec} \geq \left(\frac{\estParam\left( \muOne \right) - \estParam\left( \muTwo \right)}{2}\right)^2.
\end{equation}
\end{corollary}

\vspace{1em}

\begin{theorem}[Theorem~8 of \cite{Ma2024}, adapted to our setting]
    \label{thm:expToHighProb}
    Let $\estParamClass_0 \subseteq \estParamClass$ be non-empty. Let $\PP_0 \subseteq \PP$ such that for all $P \in \PP_0, \; \estParam(P) \in \estParamClass_0$.
    Define $D^2$ as
    \begin{equation}
        D^2 \coloneqq \sup_{P_1, P_2 \in \PP_0} \norm{\estParam\left( P_1 \right) - \estParam\left( P_2 \right)}_2^2.
    \end{equation}
    Let $\Delta \in [0, \infty)$ be such that
    \begin{equation}
        \underset{Q \in \QQ_{\eiVec}}{\inf} \; \underset{\hat{\estParam} \in \hat{\estParamClass}}{\inf} \; \underset{P \in \PP_0}{\sup} \; \E[P, Q]{\norm{\hat{\estParam}\left(Z_{1:n} \right)- \estParam \left( P \right)}_2^2} \geq \Delta.
    \end{equation}
    Then, if $D \ne 0$, for every $\varphi > 0$ and $\failureProb \in \left(0, \frac{\Delta - \varphi^2 D^2}{(1+\varphi)^2 D^2} \right)$, we have $\lowerMinimaxQ{\failureProb, \PP, \eiVec} \geq \varphi^2 D^2$.
\end{theorem}

\section{Missing Proofs from Section~\ref{sec:one-dim}} 
\newif\ifaistats
\aistatsfalse

\subsection{Proof of Theorem~\ref{thm:laplace_mech}}

\laplaceMech*

\begin{proof}
    Consider Algorithm~\ref{alg:lap}, wherein each user $i$ adds $\Lap(2/\ei)$ to their data, and the server generates an estimate, $\hat{\estParam}$, by computing a weighted sum of the users' data, weighted by $w_i \propto \ei^2$ for each user $i$.

    We are hoping for a high-probability bound on the error of this algorithm, $\abs{\hat{\estParam} - \estParam}^2.$ By the triangle inequality and a union bound, we have:
    \begin{equation}
    \label{eqn:lapErrTriangle}
        \PB{ \abs{\hat{\estParam} - \estParam}^2 > t^2 } =
        \PB{ \abs{\hat{\estParam} - \estParam} > t } \leq \PB{ \abs{\sum_{i=1}^n w_i X_i - \estParam } > \frac{t}{2}} + \PB{ \abs{\sum_{i=1}^n w_i Z_i } > \frac{t}{2}}.
    \end{equation}
    We begin by bounding the first term of \eqref{eqn:lapErrTriangle}. Observe that $\E{\sum_{i=1}^n w_i X_i} = \theta$, as $\sum_{i=1}^n w_i = 1$. Therefore, bounding the difference between $\sum_{i=1}^n w_i X_i$ and its expectation bounds the first term of \eqref{eqn:lapErrTriangle}. As $X_i \in [-1, 1]$ almost surely, $w_i X_i \in [-w_i, w_i]$ and we can apply a Hoeffding bound. This results in:
    \begin{equation}
    \label{eqn:lapErrXBound}
        \PB{ \abs{\sum_{i=1}^n w_i X_i - \estParam } > \frac{t}{2}} \leq 2 \exp \left( - \frac{t^2}{8 \sum_{i=1}^n w_i^2} \right).
    \end{equation}

    To bound the second term of \eqref{eqn:lapErrTriangle}, define $\tilde{Z}_i \coloneqq w_i Z_i$. Given that $Z_i$ is a Laplace random variable with scale $2/\ei$, $\tilde{Z}_i$ is Laplace with scale $2w_i / \ei$.   
    A standard argument from \cite{chan2011private} using the moment-generating functions of Laplace random variables shows that 
    \begin{equation}
        \E{\exp\left( \lambda \tilde{Z}_i \right)} \leq \exp \left( 8 \lambda^2 \frac{w_i^2}{\ei^2} \right) \quaaad \forall \abs{\lambda} < \frac{1}{4 \frac{w_i}{\ei}}.
    \end{equation} 
    As such, each $\tilde{Z}_i$, and consequently $\sum_i \tilde{Z}_i$, is a subexponential random variable (Definition~\ref{def:subexp}). In particular, we have that:
    \begin{equation}
        \sum_{i=1}^n \tilde{Z}_i \in \text{subE}\left( 16 \sum_{i=1}^n \frac{w_i^2}{\ei^2}, 4 \max_{i \in [n]} \frac{w_i}{\ei} \right).
    \end{equation}
    We can therefore apply Fact~\ref{fct:subE_bound} to show the concentration of the sum of $w_i Z_i$:
    \begin{equation}  
    \label{eqn:lapErrZBound}
        \PB{ \abs{ \sum_{i=1}^n w_i Z_i } > \frac{t}{2}} \leq 2 \exp \left( - \min \left\{ \frac{t^2}{128 \sum_{i=1}^n \frac{w_i^2}{\ei^2}}, \frac{t}{16 \max_{i \in [n]} \frac{w_i}{\ei} } \right\} \right).
    \end{equation}
    Combining the tail bounds (\eqref{eqn:lapErrXBound} and \eqref{eqn:lapErrZBound}) of both terms in \eqref{eqn:lapErrTriangle}, we have:
    \begin{align}
    \label{eqn:lapErrCombined}
        \PB{ \abs{\hat{\estParam} - \estParam} > t } 
        &\leq 2 \exp \left( - \frac{t^2}{8 \sum_{i=1}^n w_i^2} \right) + 2 \exp \left( - \min \left\{ \frac{t^2}{128 \sum_{i=1}^n \frac{w_i^2}{\ei^2}}, \frac{t}{16 \max_{i \in [n]} \frac{w_i}{\ei} } \right\} \right) \\
        &\leq 4 \exp \left( - \frac{1}{8} \min \left\{ \frac{t^2}{\sum_{i=1}^n w_i^2}, \frac{t^2}{16 \sum_{i=1}^n \frac{w_i^2}{\ei^2}}, \frac{t}{2 \max_{i \in [n]} \frac{w_i}{\ei} } \right\} \right). 
    \end{align}
    We can combine the first and second arguments of the minimum into one argument such that, for some constant $c$, we have:
    \begin{align}
    \label{eqn:lapErrTwoMin}
        \PB{ \abs{\hat{\estParam} - \estParam} > t } 
        &\leq 4 \exp \left( - c \min \left\{ \frac{t^2}{\sum_{i=1}^n w_i^2 \left( 1 + \frac{1}{\ei^2} \right)}, \frac{t}{2 \max_{i \in [n]} \frac{w_i}{\ei} } \right\} \right). 
    \end{align}
    Further, as $\ei \leq 1$, we know $w_i / \ei \leq w_i (1 + 1/\ei^2)$. This allows us to bring a $(1 + 1/\ei^2)$ term into the second argument, resulting in the expression:
    \begin{align}
    \label{eqn:lapErrInvolveWeights}
        \PB{ \abs{\hat{\estParam} - \estParam} > t } 
        &\leq 4 \exp \left( - c \min \left\{ \frac{t^2}{\sum_{i=1}^n w_i^2 \left( 1 + \frac{1}{\ei^2} \right)}, \frac{t}{2 \max_{i \in [n]} w_i \left( 1 + \frac{1}{\ei^2} \right)} \right\} \right). 
    \end{align}
    Substituting each $w_i$ for its realization in Algorithm~\ref{alg:lap}, we have:
    \begin{align}
    \label{eqn:lapErrNoWeights}
        \PB{ \abs{\hat{\estParam} - \estParam} > t } 
        &\leq 4 \exp \left( - c \; \sum_{i=1}^n \left( 1 + \frac{1}{\ei^2} \right)^{-1} \min \left\{ t^2, t \right\} \right). 
    \end{align}
    For $\ei \leq 1, \left( 1 + \frac{1}{\ei^2} \right)^{-1} \approx \ei^2$. Incorporating this approximation into \eqref{eqn:lapErrNoWeights} leads to:
    \begin{align}
    \label{eqn:lapErrWeights}
        \PB{ \abs{\hat{\estParam} - \estParam} > t } 
        &\leq 4 \exp \left( - c \; \sum_{i=1}^n \ei^2 \; \min \left\{ t^2, t \right\} \right). 
    \end{align}
    Altogether, this implies that with probability at least $1-\failureProb$,
    \begin{equation}
        \abs{\hat{\estParam} - \estParam}^2 \leq \max \left\{ \OO \left( \left( \frac{\log \left( 1 / \failureProb \right)}{\sum_{i=1}^n \ei^2} \right)^2 \right), \OO \left(\frac{\log \left( 1 / \failureProb \right)}{\sum_{i=1}^n \ei^2} \right) \right\}.
    \end{equation}
    When $\OO \left(\frac{\log \left( 1 / \failureProb \right)}{\sum_{i=1}^n \ei^2} \right) \leq 1$, we achieve the desired bound. If this quantity is greater than 1, outputting $\hat{\estParam} = 0$ suffices to achieve error 1. 

\end{proof}

\subsection{Proof of Theorem~\ref{thm:RR}}

\RR*

\begin{proof}
    Consider Algorithm~\ref{alg:lap}, wherein  user $i$ runs $\ei$-LDP randomized response on their data, and the server generates a estimate, $\hat{\estParam}$, by computing a weighted sum of the users' private data, weighted by $w_i \propto \ei^2$ for each user $i$.
    
    Under this construction, $\EX{\hat{\estParam}} = \EX{\sum_{i=1}^n w_i c_i Y_i} = \estParam$.
    Further, as $X_i \in \{-1, 1\}$ for all $i$, $w_i c_i Y_i~\in~\{ -w_i c_i, w_i c_i \}$. Therefore, we can apply Hoeffding's Inequality to find a high-probability bound for $\abs{\hat{\estParam} - \estParam}$:
    \begin{equation}
    \label{eqn:rrHoeffding}
        \PB{\abs{\hat{\estParam} - \estParam} \geq t} \leq 2 \exp \left(- \frac{t^2}{2 \sum_{i=1}^n w_i^2 c_i^2} \right).
    \end{equation}
    Given $w_i \propto 1/c_i^2$ and under the assumption $\ei \leq 1$ for all $i$, we have:
    \begin{equation}
    \label{eqn:rrSumOfWeights}
        \sum_{i=1}^n w_i^2 c_i^2 = \left(\sum_{i=1}^n \frac{1}{c_i^2} \right)^{-1} = \OO \left( \left( \sum_{i=1}^n \ei^2 \right)^{-1} \right).
    \end{equation}
    Combining \eqref{eqn:rrHoeffding} and \eqref{eqn:rrSumOfWeights} yields
    \begin{equation}
        \PB{\abs{\hat{\estParam} - \estParam} \geq t} \leq 2 \exp \left( \OO\left(- t^2 \sum_{i=1}^n \ei^2\right) \right),
    \end{equation}
    implying that with probability at least $1-\failureProb$, 
    \begin{equation}
        \abs{\hat{\estParam} - \estParam}^2 \leq \OO \left( \frac{\log \left( 1 / \failureProb \right) }{\sum_{i=1}^n \ei^2} \right).
    \end{equation}

    Note that if $\OO \left( \frac{\log \left( 1 / \failureProb \right) }{\sum_{i=1}^n \ei^2} \right) \geq 1$, we can simply output $\hat{\estParam} = 0$ to achieve error 1. 

\end{proof}

\subsection{Proof of Theorem~\ref{thm:lowerBound}}

\lowerBound*

\begin{proof}
    Let $\estParam \in [-1, 1]$, to be defined later. For all $i \in [n]$, define distributions $\muOne^{(i)}$ and $\muTwo^{(i)}$ constructively by drawing $X \sim \muOne^{(i)}$ or $Y \sim \muTwo^{(i)}$ such that:
    \begin{align*}
        X =
        \begin{cases}
            1 & w.p. \quad \frac{1+ \estParam }{2}\\
            -1 & w.p.  \quad \frac{1- \estParam}{2}\\
        \end{cases} 
        &&
        Y =
        \begin{cases}
            1 & w.p. \quad \frac{1 - \estParam}{2}\\
            -1 & w.p.  \quad \frac{1 +\estParam}{2}\\
        \end{cases} ~.
    \end{align*} 
    Let $\pdistOne^{(i)}$ and $\pdistTwo^{(i)}$ be the pair of distributions induced by passing $\muOne^{(i)}$ and $\muTwo^{(i)}$ through an $\ei$-differentially private channel, $C^{(i)}$. Formally, define $\pdistOne^{(i)}$ and $\pdistTwo^{(i)}$ as
    \begin{equation}
    \label{eqn:defPdistOneAndTwo}
        \pdistOne^{(i)} = C^{(i)}\left( \muOne^{(i)} \right) ~~~ \text{and} ~~~ \pdistTwo^{(i)} = C^{(i)}\left( \muTwo^{(i)} \right).
    \end{equation}
    Let $\pdistOne$ and $\pdistTwo$ be the concatenation of the $n$ distributions $\pdistOne^{(1)}, \ldots, \pdistOne^{(n)}$ and $\pdistTwo^{(1)}, \ldots, \pdistTwo^{(n)}$, respectively. Consider the KL divergence of $\pdistOne$ and $\pdistTwo$. Without loss of generality, assume that $\KL{\pdistOne}{\pdistTwo} = \min \left( \KL{\pdistOne}{\pdistTwo}, \KL{\pdistTwo}{\pdistOne}\right)$. As $\pdistOne$ and $\pdistTwo$ are product distributions, their KL divergence is the sum of the divergences of their marginals. This leads to the following bound:
    \begin{align}
        \KL{\pdistOne}{\pdistTwo} 
        &=\min \left( \KL{\pdistOne}{\pdistTwo}, \KL{\pdistTwo}{\pdistOne}\right) \\
        &\leq \frac{1}{2} \left( \KL{\pdistOne}{\pdistTwo} + \KL{\pdistTwo}{\pdistOne} \right) \\
        &= \frac{1}{2} \sum_{i=1}^n \left( \KL{\pdistOne^{(i)}}{\pdistTwo^{(i)}} + \KL{\pdistTwo^{(i)}}{\pdistOne^{(i)}}\right). \label{eqn:oneDimLBSeparatingKL}
    \end{align}
    By Theorem~\ref{thrm:KL_bound}, as the distributions induced by $\ei$-privacy channels on samples from $\muOne$ and $\muTwo$, $\pdistOne^{(i)}$ and $\pdistTwo^{(i)}$ satisfy
    \begin{equation}
    \label{eqn:duchiBound}
        \KL{\pdistOne^{(i)}}{\pdistTwo^{(i)}} + \KL{\pdistTwo^{(i)}}{\pdistOne^{(i)}} \leq \min \left( 4, e^{2\ei} \right) \cdot (e^{\ei} - 1)^2 \cdot \TVsquared{\muOne^{(i)}}{\muTwo^{(i)}}.
    \end{equation}
    By construction, $\TV{\muOne^{(i)}}{\muTwo^{(i)}} = \theta$. Combined with the fact that $\ei \leq 1$, we can bound \eqref{eqn:duchiBound} as
    \begin{equation}
    \label{eqn:duchiBoundSimple}
        \KL{\pdistOne^{(i)}}{\pdistTwo^{(i)}} + \KL{\pdistTwo^{(i)}}{\pdistOne^{(i)}} \leq 4 \cdot \left(2 \ei \right)^2 \cdot \theta^2.
    \end{equation}
    Combining this bound with \eqref{eqn:oneDimLBSeparatingKL} gives:
    \begin{equation}
        \KL{\pdistOne}{\pdistTwo} \leq 8 \theta^2 \sum_{i=1}^n \ei^2 .
    \end{equation}
    To satisfy the assumption of {Corollary}~\ref{cor:KL_bound}, choose $\theta$ as:
    \begin{equation}
        \theta^2 = \min \left( \frac{\log\left(\frac{1}{4\failureProb (1 - \failureProb)} \right)}{8 \sum_{i=1}^n \ei^2}, 1\right). 
    \end{equation}
    Then, $\KL{\pdistOne}{\pdistTwo} < \log \left( \frac{1}{4\failureProb (1- \failureProb)} \right)$ and we can apply {Corollary}~\ref{cor:KL_bound}, resulting in a bound for all $\failureProb \in (0, 1/2)$ of:
    \begin{equation}
        \lowerMinimaxQ{\failureProb, \PP, \eiVec} \geq \frac{1}{2} \abs{\theta \left(\pdistOne \right) - \theta \left( \pdistTwo \right)}^2 = \frac{1}{2} \abs{\theta+ \theta }^2 = \theta^2.
    \end{equation}
    Finally, by Theorem~\ref{thrm:quantile_bound}, we can lower bound the minimax quantile $\minimaxQ{\failureProb, \PP, \eiVec}$ by the lower minimax quantile $\lowerMinimaxQ{\failureProb, \PP, \eiVec}$. Ultimately, we have the following lower bound on $\minimaxQ{\failureProb, \PP, \eiVec}$ for $\failureProb \in (0, 1/2)$:

    \begin{equation}
        \minimaxQ{\failureProb, \PP, \eiVec} \geq \lowerMinimaxQ{\failureProb, \PP, \eiVec} \geq \theta^2 = \min\left( \frac{\log\left(\frac{1}{4\failureProb (1 - \failureProb)} \right)}{8 \sum_{i=1}^n \ei^2}, 1\right).
    \end{equation}
\end{proof}

\section{Missing Proofs from Section~\ref{sec:high-dim}} 
\newif\ifaistats
\aistatsfalse

\subsection{Proof of Theorem~\ref{thm:highDimUB}}

\highDimUB*

\begin{proof}
    Consider Algorithm~\ref{alg:highDim}, wherein each user $i$ generates an $\ei$-LDP estimate $Y_i$ of their data point $X_i$ according to Algorithm~\ref{alg:duchi}, and the server outputs a weighted average $\hat{\theta} = \sum_i w_i Y_i$. Note that Algorithm~\ref{alg:duchi} produces an unbiased estimate of $X_i$ and recall that $\E{X_i} = \estParam$. 
    Applying Cauchy-Schwarz and a union bound, we can bound the tail of this algorithm as:
    \begin{equation}
    \label{eqn:highDimUBCauchy}
        \PB{\norm{\hat{\estParam} - \estParam}_2 \leq t} \geq \PB{\norm{\sum_{i=1}^n w_i \userOut - \sum_{i=1}^n w_i X_i}_2 \leq \frac{t}{2} , \norm{\sum_{i=1}^n w_i X_i - \estParam}_2 \leq \frac{t}{2}}.
    \end{equation}
    It follows from the law of total probability that:
    \begin{equation}
    \label{eqn:highDimUBTotalProb}
        \PB{\norm{\hat{\estParam} - \estParam}_2 \leq t} \geq \PB{\norm{\sum_{i=1}^n w_i \userOut - \sum_{i=1}^n w_i X_i}_2 \leq \frac{t}{2} \mid X_1, \ldots, X_n} \cdot \PB{\norm{\sum_{i=1}^n w_i X_i - \estParam}_2 \leq \frac{t}{2}}.
    \end{equation}
    We will establish tail bounds on each of these terms separately. 
    
    \paragraph{First term of \eqref{eqn:highDimUBTotalProb}.}
    First, consider the first term. We begin by showing that, for all $i$, $\userOut - X_i$ conditioned on $X_i$ is a subgaussian random vector with variance proxy independent of $\dimValues$. Equivalently, we show that for any $\ell \in \mathbb{S}^{\dimValues-1}$, $\left\langle \ell, \userOut - X_i \right\rangle$ is subgaussian.

    Fix $i \in [n]$. Without loss of generality, let $X_i = \basisVec_\dimValues$. Let $S_1 = \{Y \in \mathbb{S}^{\dimValues - 1}(\boundi) \; \text{ s.t. } \left\langle Y, X_i \right\rangle > 0 \}$, and let $S_2 = S_1^\complement = \{Y \in \mathbb{S}^{\dimValues - 1}(\boundi) \; \text{ s.t. } \left\langle Y, X_i \right\rangle \leq 0 \}$. Let $Q_1$ be the uniform distribution over $S_1$, and let $Q_2$ be the uniform distribution over $S_2$. Denote by $Q$ the distribution of $\userOut$. Then, $Q$ is a mixture of $Q_1$ and $Q_2$. 
    
    Let $p$ be the probability that $\userOut \sim Q_1$, and let $1-p$ be the probability that $\userOut \sim Q_2$. 
    The event $\userOut \sim Q_1$ can occur in two ways: first, in Line~\ref{line:highDimAlgChoiceXTilde} of Algorithm~\ref{alg:highDim}, we choose $\tilde{X_i} \propto - X_i$ and then, in Line~\ref{line:highDimAlgChoiceY}, we have $T=0$; or, in Line~\ref{line:highDimAlgChoiceXTilde} of Algorithm~\ref{alg:highDim}, we choose $\tilde{X_i} \propto X_i$ and then, in Line~\ref{line:highDimAlgChoiceY}, we have $T=1$. 
    Given the probabilities of each of these events, we can find $p$ to be:
    \begin{equation}
    \label{eqn:highDimUBDerivationP}
        p = \frac{1}{2} + \frac{\norm{X_i}_2}{2\radius} \left( \frac{e^\ei-1}{e^\ei + 1} \right).
    \end{equation}
    Define a mirroring function $\textsc{mirror} : S_2 \rightarrow S_1$ as:
    \begin{equation}
        \textsc{mirror}\left( (x_1, \ldots, x_{\dimValues-1}, x_\dimValues) \right) = (x_1, \ldots, x_{\dimValues-1}, -x_\dimValues).
    \end{equation}
        
    Now, for an arbitrary $\ell \in \mathbb{S}^{\dimValues-1}$, let $f: \mathbb{S}^{\dimValues-1}\left( \boundi \right) \rightarrow \reals$ be such that $f(\Yt) = \langle \ell, \Yt  \rangle$. By construction, $f$ is a $1$-Lipschitz function over the sphere of radius $\boundi$. Similarly, define $g: \mathbb{S}^{\dimValues-1}\left( \boundi \right) \rightarrow \reals$ as follows:
    \begin{equation}
        g\left(\Yt\right) = \begin{cases}
            f\left( \Yt \right) & \Yt \in S_1 \\
            f\left( \textsc{mirror} \left( \Yt \right) \right) & \Yt \in S_2 
        \end{cases}.
    \end{equation}
    Because $f$ is $1$-Lipschitz, $g$ is also $1$-Lipschitz. If two points lie on the same hemisphere, their distance under $g$ is the same as under $f$, whereas if they lie on different hemispheres, the mirroring operation brings them closer together, so their distance under $g$ is less than that under $f$.

    We will now prove that the distribution of $f - \E[Q]{f}$ has subgaussian tail bounds. If $Q$ was uniform over the entire sphere, we could apply Levy's Lemma immediately to establish this result. However, $Q$ is a mixture of two uniform distributions over the sphere. To prove subgaussianity, we will show that the distribution of $f - \E[Q_1]{f}$ over $Q_1$ is subgaussian, re-express $Q$ as a mixture of the uniform distribution and $Q_1$, and prove that, because of the subgaussianity of its parts,  $f - \E[Q]{f}$ over $Q$ is subgaussian.
    
    \vspace{1ex}
    
    \noindent \textit{Subgaussianity over $Q_1$.}
    Let $\Unif$ denote the uniform distribution over $\mathbb{S}^{\dimValues-1}\left( \boundi \right)$. We begin by relating the distribution of $f - \E[Q_1]{f}$ over $S_1$ with the distribution of $g - \E[\Unif]{g}$ over $\Unif$. First, because $\Unif$ can be expressed as an equal mixture of $Q_1$ and $Q_2$, we have:
    \begin{align}
        &\PB[\Yt \sim \Unif]{\abs{g(\Yt) - \E[\Yt \sim \Unif]{g(\Yt)}}} \\
        &\quad =   \frac{1}{2} \PB[\Yt \sim Q_1]{\abs{g(\Yt) - \E[\Yt \sim \Unif]{g(\Yt)}}} + \frac{1}{2} \PB[\Yt \sim Q_2]{\abs{g(\Yt) - \E[\Yt \sim \Unif]{g(\Yt)}}} \label{eqn:highDimUBUnifMix}.
    \end{align}
    For $\Yt \sim Q_1$, we know $g(\Yt) = f(\Yt)$. Additionally, we can find the expectation of $g$ over $\Unif$ to be:
    \begin{align}
        \E[\Yt \sim \Unif]{g(\Yt)} 
        &= \frac{1}{2} \E[\Yt \sim Q_1]{g(\Yt)} + \frac{1}{2} \E[\Yt \sim Q_2]{g(\Yt)} \\
        &= \frac{1}{2} \E[\Yt \sim Q_1]{f(\Yt)} + \frac{1}{2} \E[\Yt \sim Q_2]{f(\textsc{mirror}(\Yt))} \\
        &= \frac{1}{2} \E[\Yt \sim Q_1]{f(\Yt)} + \frac{1}{2} \E[\Yt \sim Q_1]{f(\Yt)} \\
        &= \E[\Yt \sim Q_1]{f(\Yt)}.
    \end{align}
    Combining these facts with \eqref{eqn:highDimUBUnifMix}, we have:
    \begin{align}
        &\PB[\Yt \sim \Unif]{\abs{g(\Yt) - \E[\Yt \sim \Unif]{g(\Yt)}}} \\
        &\quad =  \frac{1}{2} \PB[\Yt \sim Q_1]{\abs{f(\Yt) - \E[\Yt \sim Q_1]{f(\Yt)}}} + \frac{1}{2} \PB[\Yt \sim Q_2]{\abs{g(\Yt) - \E[\Yt \sim \Unif]{g(\Yt)}}} \\
        &\quad = \frac{1}{2} \PB[\Yt \sim Q_1]{\abs{f(\Yt) - \E[\Yt \sim Q_1]{f(\Yt)}}} + \frac{1}{2} \PB[\Yt \sim Q_1]{\abs{f(\Yt) - \E[\Yt \sim \Unif]{f(\Yt)}}} \\
        &\quad = \PB[\Yt \sim Q_1]{\abs{f(\Yt) - \E[\Yt \sim Q_1]{f(\Yt)}}} \label{eqn:highDimUB_q1fromUnif}.
    \end{align}

    Because $g$ is $1$-Lipschitz over $\mathbb{S}^{\dimValues-1}\left(\boundi\right)$, we can apply Levy's Lemma (Theorem~\ref{thm:Levy}), resulting in:
    \begin{equation}
    \label{eqn:highDimUB_q1Levy}
        \PB[\Yt \sim Q_1]{\abs{f(\Yt) - \E[\Yt \sim Q_1]{f(\Yt)}}} \leq 4 \exp \left( - \frac{2 C \dimValues t^2}{\boundi^2} \right).
    \end{equation}
    Therefore, $f(\Yt) - \E[\Yt \sim Q_1]{f(\Yt)}$ is $\OO \left( \frac{\boundi^2}{\dimValues} \right)$-subgaussian over $Q_1$.
    Similarly, by Levy's Lemma (Theorem~\ref{thm:Levy}),
    \begin{equation}
    \label{eqn:highDimUB_unifLevy}
        \PB[\Yt \sim \Unif]{\abs{f(\Yt) - \E[\Yt \sim \Unif]{f(\Yt)}}} \leq 2 \exp \left( - \frac{2 C \dimValues t^2}{\boundi^2} \right),
    \end{equation}
    implying that $f(\Yt) - \E[\Yt \sim \Unif]{f(\Yt)}$ is also $\OO \left( \frac{\boundi^2}{\dimValues} \right)$-subgaussian over $\Unif$.

    \vspace{1ex}

    \noindent \textit{Subgaussianity over $Q$.}
    Recall that $Q$ is a mixture of $Q_1$ and $Q_2$:
    \begin{equation}
        Q = p \cdot Q_1 + (1-p) \cdot Q_2.
    \end{equation}
    However, because $\Unif$ is also an equal mixture of $Q_1$ and $Q_2$, we could alternatively express $Q$ as a mixture of $Q_1$ and $\Unif$ as follows:
    \begin{equation}
        Q = (2p -1) \cdot Q_1 + 2 (1-p) \cdot \Unif.
    \end{equation}
    We continue to prove that $f - \E[Q]{f}$ is subgaussian by proving that it satisfies the second property described in Definition~\ref{def:subgaussian}.

    Manipulating the expression given by the second property of Definition~\ref{def:subgaussian} for $\Unif$ and $Q_1$, respectively, we have, for some constants $c_1$ and $c_2$:
    \begin{gather}
        \E[\Yt \sim \Unif]{\exp \left( t \cdot f(\Yt) \right) } \leq \exp \left( \OO \left( \frac{\boundi^2 t^2}{\dimValues} \right) \right) \; \exp \left( t \cdot \E[Y \sim \Unif]{f(\Yt)} \right) \\
        \E[\Yt \sim Q_1]{\exp \left( t \cdot f(\Yt) \right) } \leq \exp \left( \OO \left(  \frac{\boundi^2 t^2}{\dimValues} \right) \right) \; \exp \left( t \cdot \E[Y \sim Q_1]{f(\Yt)} \right).
    \end{gather}
    Applying these bounds and the law of total expectation yields, for any $t \in \reals$,
    \begin{align}
        &\E[\Yt \sim Q]{\exp \left( t \left( f(\Yt) - \E[\Yt \sim Q]{f(\Yt)} \right) \right) } \\
        &\quad \leq \exp \left( - t \; \E[\Yt \sim Q]{f(Y)} \right) \left( 2 (1-p) \; \exp \left( \OO \left( \frac{\boundi^2 t^2}{\dimValues} \right) \right) \; \exp \left( t \cdot \E[\Yt \sim \Unif]{f(\Yt)} \right) \right. \\
        &\quaaaad \left. + (2p-1) \; \exp \left( \OO \left( \frac{\boundi^2 t^2}{\dimValues} \right) \right) \; \exp \left( t \cdot \E[\Yt \sim Q_1]{f(\Yt)} \right) \right) \label{eqn:highDimUB_MGFBound1}.
    \end{align}
    As $f$ represents an inner product between $\boundi \Yt$ and $\ell$ and the expected value of $Y$ drawn uniformly from the sphere is the zero vector, we have $\E[\Yt \sim \Unif]{f(\Yt)} = 0$. 
    Additionally, conditioned on $X_i$, the expected value of $Y$ drawn from $Q$ is $X_i$ \citep{duchi2013local}. 
    Therefore, 
    \begin{equation}
        f(X_i) = \E[\Yt \sim Q]{f(\Yt)} = (2p-1) \E[\Yt \sim Q_1]{f(\Yt)},
    \end{equation}
    implying $\E[\Yt \sim Q_1]{f(\Yt)} = \langle \ell, X_i / (2p-1) \rangle$.
    Recall that $p$ was the probability that $\userOut$ was drawn from the hemisphere on which $\langle \userOut, X_i \rangle > 0$, described in \eqref{eqn:highDimUBDerivationP}. Substituting for $p$ gives us a bound on \eqref{eqn:highDimUB_MGFBound1} of:
    \begin{align}
        &\E[\Yt \sim Q]{\exp \left( t \left( f(\Yt) - \E[\Yt \sim Q]{f(\Yt)} \right) \right) } \\ 
        &\quad \leq \exp \left( - t \; \langle \ell, X_i \rangle \right) \exp \left( \OO \left( \frac{\boundi^2 t^2}{\dimValues} \right) \right) \cdot \left(  1 +  \exp \left(  tr \left( \frac{e^\ei+1}{e^\ei - 1} \right) \frac{\left\langle \ell, X_i \right\rangle}{\norm{X_i}_2} \right) \right) \label{eqn:highDimUB_MGFBound2}.
    \end{align}
    Taking absolute values, along with the facts that $\frac{e^\ei+1}{e^\ei - 1} = \OO(1/\ei)$ and $\abs{\left\langle \ell, X_i \right\rangle / \norm{X_i}_2} \leq 1$ leaves us with:
    \begin{align*}
        \E[\Yt \sim Q]{\exp \left( t \left( f(\Yt) - \E[\Yt \sim Q]{f(\Yt)} \right) \right) } 
        &\leq \exp \left( \OO \left( \frac{\boundi^2 t^2}{\dimValues} \right) \right) \cdot \left(  \exp(\abs{t} \radius) +  \exp \left( \frac{\abs{t} \radius}{\ei} \right) \right) \\
        &\leq 2 \exp \left( \OO \left( \frac{\boundi^2 t^2}{\dimValues} \right) \right) \cdot \left( \exp \left( \frac{\abs{t} \radius}{\ei} \right) \right).
    \end{align*}
    Given $\boundi^2 \leq \radius^2 \dimValues / \ei$, this is:
    \begin{align}
        \E[\Yt \sim Q]{\exp \left( t \left( f(\Yt) - \E[\Yt \sim Q]{f(\Yt)} \right) \right) } 
        &\leq 2 \exp \left( \OO \left( \frac{\radius^2 t^2}{\ei^2} \right) \right) \cdot \left( \exp \left( \frac{\abs{t} \radius}{\ei} \right) \right) \\
        &\leq 2e \cdot \exp \left( 2 \OO \left( \frac{\radius^2 t^2}{\ei^2} \right) \right).
    \end{align}
    Therefore, by Definition~\ref{def:subgaussian}, $f(\userOut) - \E[\userOut \sim Q]{f(\userOut)} = \langle \ell, \userOut - X_i \rangle$ is $\OO \left(r^2/\ei^2 \right)$-subgaussian for all $i, \ell$. 

    \vspace{1ex}
    
    \noindent \textit{Error bound from subgaussianity.}
    
    The properties of subgaussianity and the fact that $\sum_i w_i = 1$ then imply that $\langle \ell, \sum_{i=1}^n w_i \left( \userOut - X_i \right) \rangle$ is itself subgaussian, with variance proxy at most $\OO \left( \radius^2 \sum_{i=1}^n \frac{w_i^2}{\ei^2} \right)$.
    Therefore, $\sum_{i=1}^n w_i \left( \userOut - X_i \right)$ is an $\OO \left( \radius^2 \sum_{i=1}^n \frac{w_i^2}{\ei^2} \right)$-subgaussian random vector. As such, we can apply {Fact}~\ref{fct:subGaussianNorm} to find tail bounds on its norm, therefore implying that, conditioned on $X_i, \ldots, X_n$, with probability at least $1-\failureProb / 2$,
    \begin{equation}
        \norm{\sum_{i=1}^n w_i \left( \userOut - X_i \right)}_2 \leq \OO \left( \sqrt{\radius^2 \sum_{i=1}^n \frac{w_i^2}{\ei^2} \left(d + \log \left( \frac{2}{\failureProb} \right) \right) }\right).
    \end{equation}
    Finally, as each $w_i$ is chosen proportionally to $\ei^2$, with probability at least $1 - \failureProb / 2$, conditioned on $X_i, \ldots, X_n$,
    \begin{equation}
    \label{eqn:highDimUB_boundT1}
        \norm{\sum_{i=1}^n w_i \left( \userOut - X_i \right)}_2 \leq\OO \left( \sqrt{\frac{\radius^2 \left(d + \log \left( {2}/{\failureProb} \right) \right)}{\sum_{i=1}^n {\ei^2}} } \right).
    \end{equation}

    \paragraph{Second term of \eqref{eqn:highDimUBCauchy}.}

    To bound the second term of \eqref{eqn:highDimUBCauchy}, we apply a Hoeffding-like bound for norm-subgaussian random variables \citep{jin2019}. First, note that, as $\sum_i w_i =1$, we can express this term as the norm of a sum of centered random vectors $Z_i \coloneqq w_iX_i - \E{w_i X_i}$:
    \begin{equation}
        \norm{\sum_{i=1}^n w_i X_i - \theta}_2 = \norm{\sum_{i=1}^n w_i X_i - \E{w_i X_i}}_2 = \norm{\sum_{i=1}^n Z_i}_2.
    \end{equation}
    Each of these random variables has a bounded norm. Specifically, $\norm{Z_i}_2 \leq 2 w_i \radius$. This implies that $Z_i$ is $4 w_i^2 r^2$-norm-subgaussian by {Fact}~\ref{fct:normSubGaussian} .
    Therefore, by {Fact}~\ref{fct:normSubGHoeffdingBound}, with probability at least $1 - \failureProb/2$,
    \begin{equation}
        \norm{\sum_{i=1}^n w_i X_i - \theta}_2
        = \OO \left( \sqrt{ \radius^2 \log \left( 2\dimValues / \failureProb \right) \sum_{i=1}^n w_i^2 } \right) .
    \end{equation}
    Plugging in $w_i = \ei^2 / \sum_j \epsDP_j^2$, we have, with probability at least $1-\beta/2$,
    \begin{equation}
    \label{eqn:highDimUB_boundT2}
        \norm{\sum_{i=1}^n w_i X_i - \estParam}_2 
        \leq \OO \left( \sqrt{ \frac{\radius^2 \log \left( 2\dimValues / \failureProb \right)}{ \sum_{i=1}^n \ei^2 } }\right) \leq \OO \left( \sqrt{ \frac{\radius^2 \left( \dimValues + \log \left( 2 / \failureProb \right)\right)}{ \sum_{i=1}^n \ei^2 } }\right).
    \end{equation}

    \paragraph{Both terms.}
    Setting $\frac{t}{2}$ in \eqref{eqn:highDimUBTotalProb} to be $\OO\left( \sqrt{\frac{\radius^2 \left( \dimValues + \log \left( 2 / \failureProb \right)\right)}{ \sum_{i=1}^n \ei^2 }} \right)$, we have by \eqref{eqn:highDimUB_boundT1} and \eqref{eqn:highDimUB_boundT2},
    \begin{equation}
        \PB{\norm{\sum_{i=1}^n w_i \userOut - \sum_{i=1}^n w_i X_i}_2 \leq \frac{t}{2} \mid X_1, \ldots, X_n} \cdot \PB{\norm{\sum_{i=1}^n w_i X_i - \estParam}_2 \leq \frac{t}{2}} \geq \left(1-\frac{\failureProb}{2} \right)^2 \geq 1 - \failureProb.
    \end{equation}
    Combined with \eqref{eqn:highDimUBTotalProb}, this implies that with probability at least $1 - \failureProb$,
    \begin{equation}
    \label{eqn:highDimUB_boundT1andT2}
        \norm{\hat{\estParam} - \estParam}_2 
        \leq \OO \left( \sqrt{ \frac{\radius^2 \left( \dimValues  + \log \left( 1 / \failureProb \right)\right)}{ \sum_{i=1}^n \ei^2 } } \right) .
    \end{equation}
    An upper bound of $\radius^2$ can be achieved by outputting the zero vector.
    
\end{proof}

\subsection{Proof of Theorem~\ref{thm:highDimLB}}

\highDimLB*

\begin{proof}
    By Theorem~\ref{thm:lowerBound}, we have
    \begin{equation}
        \minimaxQ{\failureProb, \PP_{2, \radius}, \eiVec} \geq \radius^2 \min \left( \log\left(\frac{1}{\failureProb}\right), 1 \right).
    \end{equation} 
    
    To establish the remaining terms, we begin by proving a lower bound on the minimax expected error of the multi-dimensional mean estimation problem.  
    Specifically, for all $k \in [d]$, we will construct  $\PP^{(k)} \subseteq \PP_{2, \radius}$ such that:
    \begin{equation}
        \underset{Q \in \QQ_{\eiVec}}{\inf} \; \underset{\hat{\estParam} \in \hat{\estParamClass}}{\inf} \; \underset{P_\nu \in \PP^{(k)} }{\sup} \; \E[P_\nu, Q]{\norm{\hat{\estParam}\left(Z_{1:n} \right)- \estParam \left( P_\nu \right)}_2^2} \geq \Theta\left( \radius^2 \min\left(\frac{1}{k}, \frac{k}{\sum_{i=1}^n \ei^2} \right) \right).
    \end{equation}
    To do so, let $\psi \in (0, 1]$, to be specified later in the proof. Let $k \in [\dimValues]$. Construct $\PP^{(k)} \subseteq \PP_{2, \radius}$ as follows. Let $\VV_k = \{ -1, 1\}^k$. For all $\nu \in \VV_k$ and for all $i \in [n]$, define a distribution $\muNuI \in \PP^{(k)}$ constructively by first drawing $j \sim \Unif\left( [k] \right)$, then drawing $X$ such that
    \begin{equation*}
        X = \begin{cases}
            r \basisVec_j & \text{w.p.} \frac{1+\psi \nu_j}{2} \\
            -r \basisVec_j & \text{w.p.} \frac{1-\psi \nu_j}{2} 
        \end{cases},
    \end{equation*}
    where $\basisVec_j$ is the $j^{\text{th}}$ basis vector in $\reals^\dimValues$.
    For all $\nu \in \VV_k$, let $\pdistNuI$ be the distribution induced by passing $\muNuI$ through an $\ei$-differentially private channel, $C^{(i)}$. Formally, $\pdistNuI = C^{(i)}\left( \muNuI \right)$. Let $\pdistNu$ be the concatenation of the $n$ distributions $\pdistNu^{(1)}, \ldots, \pdistNu^{(n)}$. For all $\nu, \nu' \in \VV_k$, we write $\nu \sim \nu'$ if $\nu$ and $\nu'$ differ in exactly one coordinate. If $\nu$ and $\nu'$ differ only in coordinate $j$, we write $\nu \sim_j \nu'$.
    
    Then, under Assouad's Lemma (Lemma~\ref{lem:assouad}), we have
    \begin{align}
        \underset{Q \in \QQ_{\eiVec}}{\inf} \; \underset{\hat{\estParam} \in \hat{\estParamClass}}{\inf} \; \underset{P_\nu \in \PP^{(k)}}{\sup} \; &\E[P_\nu, Q]{\sum_{j=1}^k \left(\hat{\estParam}\left(Z_{1:n} \right)_j - \estParam \left( P_\nu \right)_j \right)^2} \nonumber \\
        &\geq  \frac{k}{2} \; \min_{j \in [k], \nu \sim_j \nu'} \left( \estParam\left( \pdistNu \right)_{j} - \estParam\left( \pdistNuTwo \right)_{j} \right)^2 \; \min_{\nu \sim \nu'} \left( 1 - \TV{Q_\nu}{Q_{\nu'}} \right) \label{eqn:assouadWithTV}.
    \end{align}
    Consider the total variation distance between $\pdistNu$ and $\pdistNuTwo$ for $\nu \sim \nu'$. By Pinsker's Inequality, we can bound the square of this total variation distance by the KL-divergence of $\pdistNu$ and $\pdistNuTwo$ as follows:
    \begin{align}
        \TVsquared{\pdistNu}{\pdistNuTwo} &\leq \frac{1}{2} \min \left( \KL{\pdistNu}{\pdistNuTwo}, \KL{\pdistNuTwo}{\pdistNu} \right) \\
        &\leq \frac{1}{4}  \left( \KL{\pdistNu}{\pdistNuTwo} + \KL{\pdistNuTwo}{\pdistNu} \right) \\
        &= \frac{1}{4} \sum_{i=1}^n \left( \KL{\pdistNuI}{\pdistNuTwoI} + \KL{\pdistNuTwoI}{\pdistNuI} \right) \label{eqn:assouadPinsker},
    \end{align}
    where the equality is due to $\pdistNu$ and $\pdistNuTwo$ being product distributions. By Theorem~\ref{thrm:KL_bound}, as the distributions induced by $\ei$-privacy channels on samples from $\muNuI$ and $\muNuTwoI$, $\pdistNuI$ and $\pdistNuTwoI$ satisfy:
    \begin{equation}
    \label{eqn:assouadDataProc}
        \KL{\pdistNuI}{\pdistNuTwoI} + \KL{\pdistNuTwoI}{\pdistNuI} \leq \min \left( 4, e^{2\ei} \right) \cdot (e^{\ei} - 1)^2 \cdot \TVsquared{\muNuI}{\muNuTwoI}.
    \end{equation}
    By the construction of $\muNuI$ and $\muNuTwoI$ and given that $\nu$ and $\nu'$ differ in exactly one coordinate, $\TV{\muNuI}{\muNuTwoI} = \psi / k$. Combined with the fact that $\ei \leq 1$, we can bound \eqref{eqn:assouadDataProc} as
    \begin{equation}
        \KL{\pdistNuI}{\pdistNuTwoI} + \KL{\pdistNuTwoI}{\pdistNuI} \leq  \frac{4 \cdot \left( 2\ei \right)^2 \cdot \psi^2}{k^2}.
    \end{equation}
    Combining this bound with \eqref{eqn:assouadPinsker} gives:
    \begin{equation}
        \TVsquared{\pdistNu}{\pdistNuTwo} \leq 4 \frac{\psi^2}{k^2} \sum_{i=1}^n \ei^2 .
    \end{equation}
    Choose $\psi^2 = \min \left( k^2 / \left(16 \sum_{i=1}^n \ei^2\right), 1\right)$. Then, $\TV{\pdistNu}{\pdistNuTwo} \leq 1/2$. As a result, we can lower bound \eqref{eqn:assouadWithTV} as:
    \begin{align}
        \underset{Q \in \QQ_{\eiVec}}{\inf} \; \underset{\hat{\estParam} \in \hat{\estParamClass}}{\inf} \; \underset{P_\nu \in \PP^{(k)}}{\sup} \; \E[P_\nu, Q]{\sum_{j=1}^k \left(\hat{\estParam}\left(Z_{1:n} \right)_j - \estParam \left( P_\nu \right)_j \right)^2} 
        \geq  \frac{k}{4} \; \min_{j \in [k], \nu \sim_j \nu'} \left( \estParam\left( \pdistNu \right)_{j} - \estParam\left( \pdistNuTwo \right)_{j} \right)^2 \label{eqn:assouadNoTV},
    \end{align}
    Consider now, for all $j \in [k]$, $\min_{\nu \sim_j \nu'} \left( \estParam\left( \pdistNu \right)_{j} - \estParam\left( \pdistNuTwo \right)_{j} \right)^2$. 
    For all $\nu$, by the construction of $\pdistNu$, 
    $\estParam \left(\pdistNu \right)_j = \E{\pdistNu}_j = \frac{\psi \radius}{k} \nu_j$. Given $\pdistNu$ and $\pdistNuTwo$ such that $\nu \sim_j \nu'$, we have:
    \begin{equation}
         \left( \estParam\left( \pdistNu \right)_{j} - \estParam\left( \pdistNuTwo \right)_{j} \right)^2 = \left\langle \basisVec_j, \theta_{\nu, j} - \theta_{\nu', j} \right\rangle^2 = \frac{\psi^2 r^2}{k^2} \left\langle \basisVec_j, \nu - \nu' \right\rangle^2 = \frac{4 \psi^2 r^2}{k^2}. 
    \end{equation}
    Substituting this equality into \eqref{eqn:assouadNoTV} leads to:
    \begin{equation}
        \underset{Q \in \QQ_{\eiVec}}{\inf} \; \underset{\hat{\estParam} \in \hat{\estParamClass}}{\inf} \; \underset{P_\nu \in \PP^{(k)}}{\sup} \; \E[P_\nu, Q]{\sum_{j=1}^k \left(\hat{\estParam}\left(Z_{1:n} \right)_j - \estParam \left( P_\nu \right)_j \right)^2} 
        \geq \frac{ \psi^2 \radius^2}{k}.
        \label{eqn:AssouadExpBoundSumToK}
    \end{equation}
    
    Ultimately, as the sum inside the expectation of \eqref{eqn:AssouadExpBoundSumToK} is upper bounded by the $\ell_2^2$-norm of $\hat{\estParam} - \estParam(P_\nu)$, we have a lower bound on the minimax risk for the class $\PP^{(k)}$, given by:
    \begin{equation}
        \underset{Q \in \QQ_{\eiVec}}{\inf} \; \underset{\hat{\estParam} \in \hat{\estParamClass}}{\inf} \; \underset{P_\nu \in \PP^{(k)}}{\sup} \; \E[P_\nu, Q]{\norm{\hat{\estParam}\left(Z_{1:n} \right) - \estParam \left( P_\nu \right)}_2^2} 
        \geq \underset{Q \in \QQ_{\eiVec}}{\inf} \; \underset{\hat{\estParam} \in \hat{\estParamClass}}{\inf} \; \underset{P_\nu \in \PP^{(k)}}{\sup} \; \E[P_\nu, Q]{\sum_{j=1}^k \left(\hat{\estParam}\left(Z_{1:n} \right)_j - \estParam \left( P_\nu \right)_j \right)^2} 
        \geq \frac{ \psi^2 \radius^2}{k}.
        \label{eqn:AssouadExpBoundNom}
    \end{equation}

    To translate this local minimax risk bound into a bound on the related minimax quantile, we will apply Theorem~\ref{thm:expToHighProb}. Let $D^2 = \sup_{P_\nu, P_{\nu'} \in \PP^{(k)}} \norm{\estParam\left( P_\nu \right) - \estParam\left( P_{\nu'} \right)}^2$. We can calculate $D^2$ as:
    \begin{equation}
        D^2 = \underset{P_\nu, P_{\nu'} \in \PP^{(k)}}{\sup} \norm{ \theta \left( \muNu \right) - \theta  \left( \muNuTwo \right)}_2^2 = \frac{\psi^2 \radius^2}{k^2} \underset{\nu, \nu' \in \VV_k}{\sup} \norm{\nu - \nu'}_2^2 = \frac{4 \psi^2 \radius^2}{k} 
    \end{equation}
    Therefore, by Theorem~\ref{thm:expToHighProb} with $\varphi = 1/4$, we have, for all $\failureProb \in \left(0, \frac{3}{20}\right)$, 
    \begin{equation}
        \lowerMinimaxQ{\failureProb, \PP_{2, \radius}, \eiVec} 
        \geq \frac{\psi^2 \radius^2}{4 k}.
    \end{equation}
    By Theorem~\ref{thrm:quantile_bound}, we can lower bound the minimax quantile $\minimaxQ{\failureProb, \PP_{2,\radius}, \eiVec}$ by the lower minimax quantile $\lowerMinimaxQ{\failureProb, \PP_{2,\radius}, \eiVec}$, yielding:
    \begin{equation}
        \minimaxQ{\failureProb, \PP_{2,\radius}, \eiVec} \geq \lowerMinimaxQ{\failureProb, \PP_{2,\radius}, \eiVec} \geq  \frac{\psi^2 \radius^2}{4 k}.
    \end{equation}
    Substituting for $\psi^2 = \min\left( 1, \frac{k^2}{16 \sum_{i=1}^n \ei^2} \right)$, we have
    \begin{equation}
        \label{eqn:highDimLBNoMaxK}
        \minimaxQ{\failureProb, \PP_{2,\radius}, \eiVec} \geq \frac{\radius^2}{4}  \min\left(\frac{1}{k}, \frac{k}{16 \sum_{i=1}^n \ei^2} \right).
    \end{equation}
    Given that \eqref{eqn:highDimLBNoMaxK} holds for all $k \in [\dimValues]$, we can lower bound $\minimaxQ{\failureProb, \PP_{2,\radius}, \eiVec}$ by a maximization over all $k$ as follows:
    \begin{equation}
        \label{eqn:highDimLBAssouadPart}
        \minimaxQ{\failureProb, \PP_{2,\radius}, \eiVec} \geq \frac{\radius^2}{4} \max_{k \in [\dimValues]} \; \min\left(\frac{1}{k}, \frac{k}{16 \sum_{i=1}^n \ei^2} \right) = \frac{\radius^2}{4} \min\left(1, \frac{d}{16 \sum_{i=1}^n \ei^2}, \frac{1}{4 \sqrt{\sum_{i=1}^n \ei^2}} \right).
    \end{equation}
    Finally, combining Theorem~\ref{thm:lowerBound} and \eqref{eqn:highDimLBAssouadPart}, we find, for some constant $c_2 > 0$,
    \begin{equation}
    \label{eqn:highDimLBFinal}
        \minimaxQ{\failureProb, \PP_{2,\radius}, \eiVec} \geq c_2 \cdot \radius^2 \min \left(1, \frac{\log \left( {1}/{\failureProb} \right) +  \dimValues}{\sum_{i=1}^n \ei^2}, \frac{\log \left( {1}/{\failureProb} \right)}{\sum_{i=1}^n \ei^2} + \frac{1}{{\sqrt{\sum_{i=1}^n \ei^2}}} \right).
    \end{equation}
\end{proof}

\section{Missing Proofs from Section~\ref{sec:hist-est}} 
\newif\ifaistats
\aistatsfalse


\subsection{Proof of Theorem~\ref{thm:HistEstBassilyUB}}
\HistEstBassilyUB*

The proof of Theorem~\ref{thm:HistEstBassilyUB} relies on the following claim.

\begin{restatable}{lem}{claimTwoSix}
\label{lemma:2.6}
    Let $\eiVec = \eis$, with $\ei \leq 1$ for all $i$. Let $\failureProb \in (0, 1)$. Fix $\projMatrix \in \{-{1}{\sqrt{\dimCompressed}}, {1}{\sqrt{\dimCompressed}}\}^{\dimCompressed \times \dimValues}$. Let $\trueValue \in [\dimValues]$. Let $\compressedMean$ be the result of Line~\ref{line:compressedMean} of Algorithm~\ref{alg:histEstBassily}. Then, for $\trueValue \in [\dimValues]$, $\compressedMean$ satisfies:
    \begin{equation}
        \PB{\abs{\left\langle \compressedMean - \E{\compressedMean}, \projMatrix \basisVec_\trueValue \right\rangle} \leq \sqrt{2 \log \left( \frac{2}{\failureProb} \right) \cdot \left( \sum_{i=1}^n \frac{1}{c_i^2} \right)^{-1}}} \geq 1 - \failureProb,
    \end{equation}
    where the probability is taken over the randomness in sampling $j$ and $\privateCompVecI$ for each user.
\end{restatable}

\begin{proof}[Proof of Claim~\ref{lemma:2.6}]
    Rewriting $\compressedMean$ as $\sum_i w_i \privateCompVecI$, we have:
    \begin{align*}
        \abs{\left\langle \compressedMean - \E{\compressedMean}, \projMatrix \basisVec_\trueValue \right\rangle}
        &= \abs{\left\langle \sum_{i=1}^n w_i \privateCompVecI - \E{\sum_{i=1}^n w_i \privateCompVecI}, \projMatrix \basisVec_\trueValue \right\rangle} \\
        &= \abs{\left\langle \sum_{i=1}^n w_i \privateCompVecI, \projMatrix \basisVec_\trueValue \right\rangle - \left\langle \E{\sum_{i=1}^n w_i \privateCompVecI}, \projMatrix \basisVec_\trueValue \right\rangle} \\
        &= \abs{\sum_{i=1}^n w_i \left\langle \privateCompVecI, \projMatrix \basisVec_\trueValue \right\rangle - \E{\sum_{i=1}^n w_i \left\langle \privateCompVecI, \projMatrix \basisVec_\trueValue \right\rangle}}.
    \end{align*}
    We have that, for each $i$, 
    \begin{equation*}
        \left\langle \privateCompVecI, \projMatrix \basisVec_\trueValue \right\rangle = \sum_{j=1}^\dimValues \privateCompVecI{}_{j} \projMatrix_{j\trueValue}.
    \end{equation*}
    Each $\privateCompVecI$ is non-zero for one index $j' \in [\dimValues]$. This implies that $\left\langle \privateCompVecI, \projMatrix \basisVec_\trueValue \right\rangle = \privateCompVecI{}_{j'} \projMatrix_{j'\trueValue}$.
    By construction, $\privateCompVecI \in \{ -c_i \sqrt{\dimValues}, 0, c_i \sqrt{\dimValues}\}^\dimCompressed$ for all $i$ and $\projMatrix \in \{-1/\sqrt{\dimCompressed}, 1/\sqrt{\dimCompressed}\}^{\dimCompressed \times \dimValues}$. As a result, we know that $\left\langle \privateCompVecI, \projMatrix \basisVec_\trueValue \right\rangle \in \{-c_i, c_i\}$ and $w_i \left\langle \privateCompVecI, \projMatrix \basisVec_\trueValue \right\rangle \in \{-w_i c_i, w_i c_i\}$. We can thus apply a standard Hoeffding bound and observe that, for all $t>0$,
    \begin{equation}
    \label{eqn:histEstHoeffdingBound}
        \PB{\abs{\sum_{i=1}^n w_i \left\langle \privateCompVecI, \projMatrix \basisVec_\trueValue \right\rangle - \E{\sum_{i=1}^n w_i \left\langle \privateCompVecI, \projMatrix \basisVec_\trueValue \right\rangle}} \geq t} \leq 2\exp\left( - \frac{t^2}{2 \sum_{i=1}^n w_i^2 c_i^2} \right).
    \end{equation}
    As each $w_i$ is chosen proportionally to $1/c_i^2$, \eqref{eqn:histEstHoeffdingBound}
    is: 
    \begin{equation}
    \label{eqn:histEstBound}
        \PB{\abs{\sum_{i=1}^n w_i \left\langle \privateCompVecI, \projMatrix \basisVec_\trueValue \right\rangle - \E{\sum_{i=1}^n w_i \left\langle \privateCompVecI, \projMatrix \basisVec_\trueValue \right\rangle}} \geq t} \leq 2\exp\left( - \frac{t^2}{2} \sum_{i=1}^n \frac{1}{c_i^2} \right).
    \end{equation}
    Setting $t^2 = 2 \log \left( \frac{2}{\failureProb} \right) \cdot \left( \sum_{i=1}^n \frac{1}{c_i^2} \right)^{-1}$ in \eqref{eqn:histEstHoeffdingBound} completes the proof.
\end{proof}

Given Claim~\ref{lemma:2.6}, we can now prove Theorem~\ref{thm:HistEstBassilyUB}.

\begin{proof}[Proof of Theorem~\ref{thm:HistEstBassilyUB}]
For all values $\trueValue \in [\dimValues]$, let $p(\trueValue)$ be the true probability of observing $\trueValue$ under $P$ and let $\hat{p}(\trueValue)$ be Algorithm~\ref{alg:histEstBassily}'s estimate of that probability given the observed values $\obsValue_1, \ldots, \obsValue_n$. By definition, \begin{equation}
    \norm{\hat{P}\left( \obsValue_{1:n} \right) - P}_\infty = \max_{\trueValue \in [\dimValues]} \abs{\hat{p}(\trueValue) - p(\trueValue)}.
\end{equation}
Our algorithm constructs $\hat{p}(\trueValue)$ as
$\hat{p}(\trueValue) = \left\langle \compressedMean, \projMatrix \basisVec_\trueValue \right\rangle$. As $\sum_i w_i = 1$, we can rewrite $p(\trueValue)$ as $\E[\vecValues]{ \left\langle  \sum_i w_i \basisVec_{\vecValuesI}, \basisVec_\trueValue \right\rangle }$. This leads us to:
\begin{align}
    \max_{\trueValue \in [\dimValues]} \abs{\hat{p}(\trueValue) - p(\trueValue)} &= \max_{\trueValue \in [\dimValues]} \abs{ \left\langle \compressedMean, \projMatrix \basisVec_\trueValue \right\rangle - \E[\vecValues]{ \left\langle  \sum_{i=1}^n w_i \basisVec_{\vecValuesI}, \basisVec_\trueValue \right\rangle } } \nonumber \\
    &= \max_{\trueValue \in [\dimValues]} \abs{ \left\langle \compressedMean - \E{\compressedMean} + \E{\compressedMean}, \projMatrix \basisVec_\trueValue \right\rangle - \E[\vecValues]{ \left\langle  \sum_{i=1}^n w_i \basisVec_{\vecValuesI}, \basisVec_\trueValue \right\rangle } } \nonumber \\
    &= \max_{\trueValue \in [\dimValues]} \abs{ \left\langle \compressedMean - \E{\compressedMean}, \projMatrix \basisVec_\trueValue \right\rangle + \left\langle \E{\compressedMean}, \projMatrix \basisVec_\trueValue \right\rangle - \E[\vecValues]{ \left\langle  \sum_{i=1}^n w_i \basisVec_{\vecValuesI}, \basisVec_\trueValue \right\rangle } } \nonumber.
\end{align}
We can separate these terms by applying Cauchy-Schwarz:
\begin{align}
    \max_{\trueValue \in [\dimValues]} \abs{\hat{p}(\trueValue) - p(\trueValue)}
    &\leq \max_{\trueValue \in [\dimValues]} \abs{ \left\langle \compressedMean - \E{\compressedMean}, \projMatrix \basisVec_\trueValue \right\rangle} + \max_{\trueValue \in [\dimValues]} \abs{\left\langle \E{\compressedMean}, \projMatrix \basisVec_\trueValue \right\rangle - \E[\vecValues]{ \left\langle  \sum_{i=1}^n w_i \basisVec_{\trueValue}, \basisVec_\trueValue \right\rangle } } \label{eqn:histEstErrorCauchySchwarz}.
\end{align}
Applying Claim~\ref{lemma:2.6} and a union bound, with probability at least $1 - \failureProb/2$, the first term of Equation~\eqref{eqn:histEstErrorCauchySchwarz} satisfies, for all $\trueValue \in [\dimValues]$ 
\begin{equation}
\label{eqn:histEstErrorTerm1}
    \max_{\trueValue \in [\dimValues]} \abs{ \left\langle \compressedMean - \E{\compressedMean}, \projMatrix \basisVec_\trueValue \right\rangle} \leq \sqrt{2 \log \left(\frac{4\dimValues}{\failureProb} \right) \cdot \left( \sum_{i=1}^n \frac{1}{c_i^2} \right)^{-1} }.
\end{equation}
To bound the second term, we can express $\E{\compressedMean}$ as $\E[\vecValues]{\E{\compressedMean \mid \vecValues}}$, where $\vecValues \in [\dimValues]^n$ is the vector of values observed by the algorithm. Conditioned on knowing $\vecValues$, the expected value of $\compressedMean$ is:
\begin{equation*}
    \E{\compressedMean \mid \vecValues} = \E{\sum_{i=1}^n w_i z_i \mid \vecValues} = \projMatrix \sum_{i=1}^n w_i \basisVec_{\vecValuesI}.
\end{equation*}
Therefore, the second term of Equation~\eqref{eqn:histEstErrorCauchySchwarz} is:
\begin{align*}
    \max_{\trueValue \in [\dimValues]} &\abs{\left\langle \E{\compressedMean}, \projMatrix \basisVec_\trueValue \right\rangle - \E[\vecValues]{ \left\langle  \sum_{i=1}^n w_i \basisVec_{\vecValuesI}, \basisVec_\trueValue \right\rangle } } \\
    &= \max_{\trueValue \in [\dimValues]} \abs{\left\langle \E[\vecValues]{\projMatrix \sum_{i=1}^n w_i \basisVec_{\vecValuesI}}, \projMatrix \basisVec_\trueValue \right\rangle - \E[\vecValues]{ \left\langle  \sum_{i=1}^n w_i \basisVec_{\vecValuesI}, \basisVec_\trueValue \right\rangle } } \\
    &= \max_{\trueValue \in [\dimValues]} \abs{\E[\vecValues]{\left\langle \projMatrix \sum_{i=1}^n w_i \basisVec_{\vecValuesI}, \projMatrix \basisVec_\trueValue \right\rangle - \left\langle  \sum_{i=1}^n w_i \basisVec_{\vecValuesI}, \basisVec_\trueValue \right\rangle } }.
\end{align*}
By the Johnson-Lindenstrauss lemma (Lemma~\ref{lem:JL}), with probability at least $1- \failureProb/2$,
\begin{align}
    \max_{\trueValue \in [\dimValues]} \abs{\E[\vecValues]{\left\langle \projMatrix \sum_{i=1}^n w_i \basisVec_{\vecValuesI}, \projMatrix \basisVec_\trueValue \right\rangle - \left\langle  \sum_{i=1}^n w_i \basisVec_{\vecValuesI}, \basisVec_\trueValue \right\rangle } }
    &\leq \max_{\trueValue \in [\dimValues]} \abs{ \JLconstant \cdot \E[\vecValues]{ \OO\left( \norm{\sum_{i=1}^n w_i \basisVec_{\vecValuesI}}^2 + \norm{\basisVec_\trueValue}^2 \right) } } \nonumber \\
    &= \OO(\JLconstant). \label{eqn:histEstErrorTerm2}
\end{align}
Recall that we choose 
\begin{equation*}
    \JLconstant = \OO\left( \sqrt{\log \left( \frac{\dimValues}{\failureProb} \right) \left( \sum_{i=1}^n \frac{1}{c_i^2} \right)^{-1}} \right).
\end{equation*}
Combining Equation~\eqref{eqn:histEstErrorTerm1} and Equation~\eqref{eqn:histEstErrorTerm2}, along with this choice of $\JLconstant$ and the fact that $\frac{1}{c_i^2}~=~\OO \left( \ei^2 \right)$ for all $i$, the following bound on the error of our mechanism for histogram estimation holds with probability at least $1-\beta$:
\begin{equation}
    \max_{\trueValue \in [\dimValues]} \abs{\hat{p}(\trueValue) - p(\trueValue)} \leq \OO\left(\sqrt{\frac{\log \left({\dimValues} / {\failureProb} \right)}{\sum_{i=1}^n \ei^2 }} \right).
\end{equation}
The upper bound of 1 on the error described in the theorem statement follows by outputting $\hat{p}(\trueValue) = 0$ for all $\trueValue$ but one $\trueValue'$, for which we output $\hat{p}(\trueValue') = 1$. 
\end{proof}

\end{document}